\pdfoutput=1

\documentclass[11pt]{article}

\usepackage[final]{acl}
\usepackage{xspace}

\usepackage{times}
\usepackage[most]{tcolorbox}
\usepackage{latexsym}
\usepackage[main=english,ukrainian]{babel}
\usepackage{booktabs}
\usepackage{multirow}
\usepackage{hhline}
\usepackage{makecell}

\usepackage{graphicx}
\usepackage{tabularx}
\usepackage{array}
\usepackage{hyperref}
\usepackage{natbib}
\usepackage{booktabs}
\usepackage{color, colortbl}
\usepackage{enumitem}
\usepackage{caption}
\usepackage{adjustbox}
\usepackage{amssymb}
\usepackage{utfsym}
\usepackage{mathtools}
\usepackage{amsmath}
\usepackage{inconsolata}
\usepackage{subcaption}
\usepackage{enumitem}
\usepackage{xcolor}
\usepackage[most]{tcolorbox}
\usepackage{textcomp}

\usepackage[utf8]{inputenc}
\usepackage{url}

\usepackage{microtype}

\usepackage{inconsolata}

\definecolor{lightblue}{RGB}{220, 240, 255}

\definecolor{LightYellow}{RGB}{255, 255, 204}
\definecolor{LightOrange}{RGB}{255, 229, 204}
\definecolor{LightPurple}{RGB}{230, 230, 250}
\definecolor{LightBlue}{RGB}{220, 235, 255}
\definecolor{LightGreen}{RGB}{220, 255, 220}
\definecolor{LightRed}{RGB}{255, 220, 220}

\newcommand{\emo}{\textsc{EmoBench-UA}\xspace}

\title{\emo: A Benchmark Dataset \\ for Emotion Detection in Ukrainian}

\author{Daryna Dementieva\textsuperscript{1,3}, Nikolay Babakov\textsuperscript{2} \and {\bf Alexander Fraser}\textsuperscript{1,3,4} \\
\textsuperscript{1}Technical University of Munich (TUM), \textsuperscript{2}Universidade de Santiago de Compostela, \\
\textsuperscript{3}Munich Center for Machine Learning (MCML), $^{4}$Munich Data Science Institute \\
\href{mailto:daryna.dementieva@tum.de}{\texttt{\small daryna.dementieva@tum.de}},
\href{mailto:nikolay.babakov@usc.es}{\texttt{\small nikolay.babakov@usc.es}}
}

\begin{document}
\maketitle
\begin{abstract}
While Ukrainian NLP has seen progress in many texts processing tasks, emotion classification remains an underexplored area with no publicly available benchmark to date. In this work, we introduce \emo, the first annotated dataset for emotion detection in Ukrainian texts. Our annotation schema is adapted from the previous English-centric works on emotion detection~\cite{mohammad-etal-2018-semeval,mohammad-2022-ethics-sheet} guidelines. The dataset was created through crowdsourcing using the Toloka.ai platform ensuring high-quality of the annotation process. Then, we evaluate a range of approaches on the collected dataset, starting from linguistic-based baselines, synthetic data translated from English, to large language models (LLMs). Our findings highlight the challenges of emotion classification in non-mainstream languages like Ukrainian and emphasize the need for further development of Ukrainian-specific models and training resources.
\end{abstract}

\section{Introduction}
Recent trends in natural language processing indicate a shift from predominantly monolingual English-centric approaches toward more inclusive multilingual solutions that support less-resourced and non-mainstream languages. Although cross-lingual transfer techniques---such as Adapter modules~\cite{pfeiffer-etal-2020-adapterhub} or translation from resource-rich languages~\cite{DBLP:journals/apin/KumarPR23}---have shown promise, the development of high-quality, language-specific datasets remains essential for achieving robust and culturally accurate performance in these settings.

\begin{figure}[ht!]
    \centering
    \includegraphics[width=0.9\linewidth]{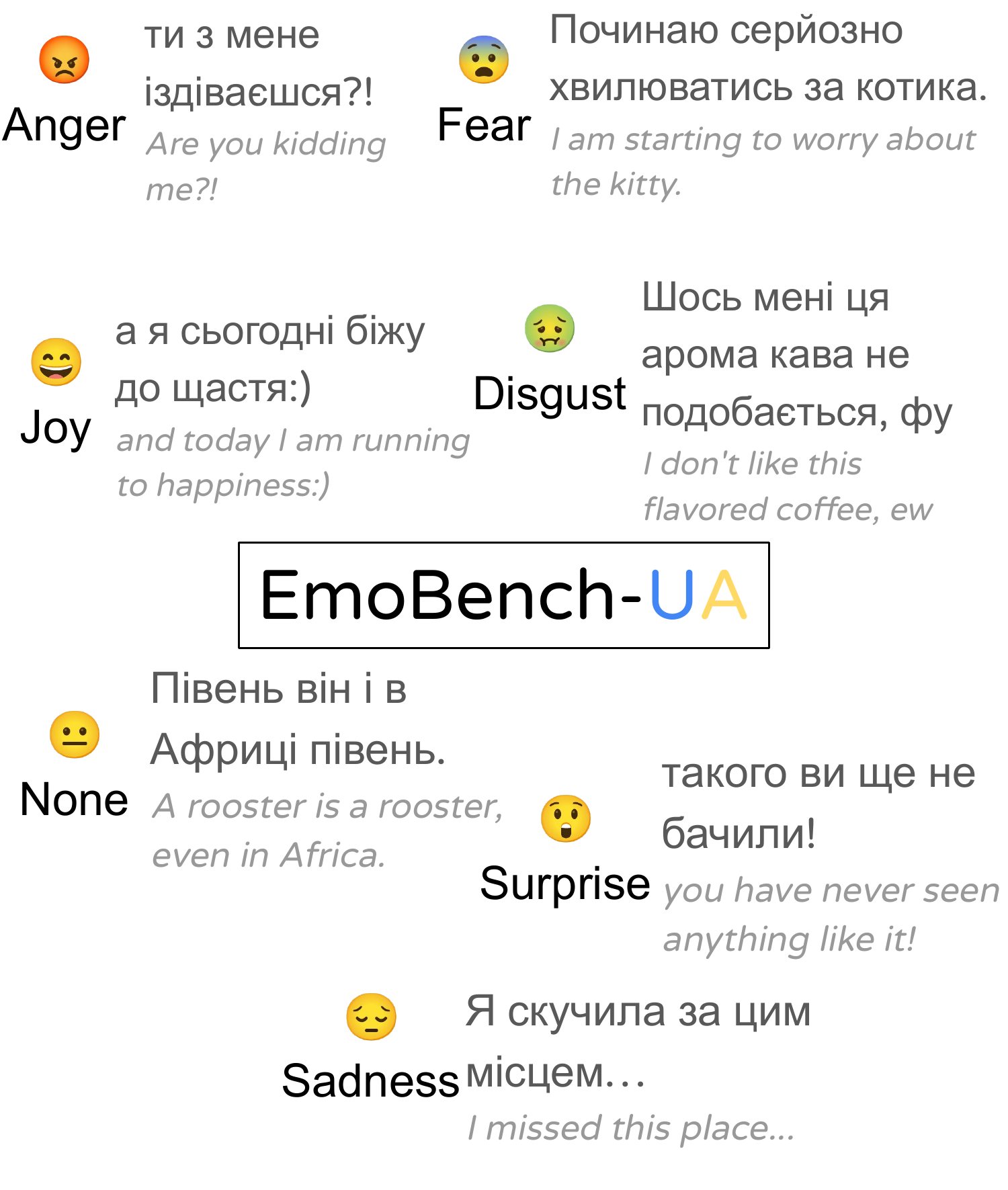}
    \caption{\emo is a benchmark of basic emotions---Joy, Anger, Fear, Disgust, Surprise, Sadness, or None---detection in Ukrainian texts.}
    \label{fig:intro}
\end{figure}

For the Ukrainian language, significant progress has been made in the development of resources for various stylistic classification tasks, such as sentiment analysis~\cite{DBLP:conf/colins/ZalutskaMSMPBK23} and toxicity detection~\cite{dementieva-etal-2024-toxicity}. However, to the best of our knowledge, no publicly available dataset has yet addressed the task of emotion classification. In this work, we aim to fill this gap through the following contributions:

\begin{figure*}[ht!]
    \centering
    \includegraphics[width=0.85\linewidth]{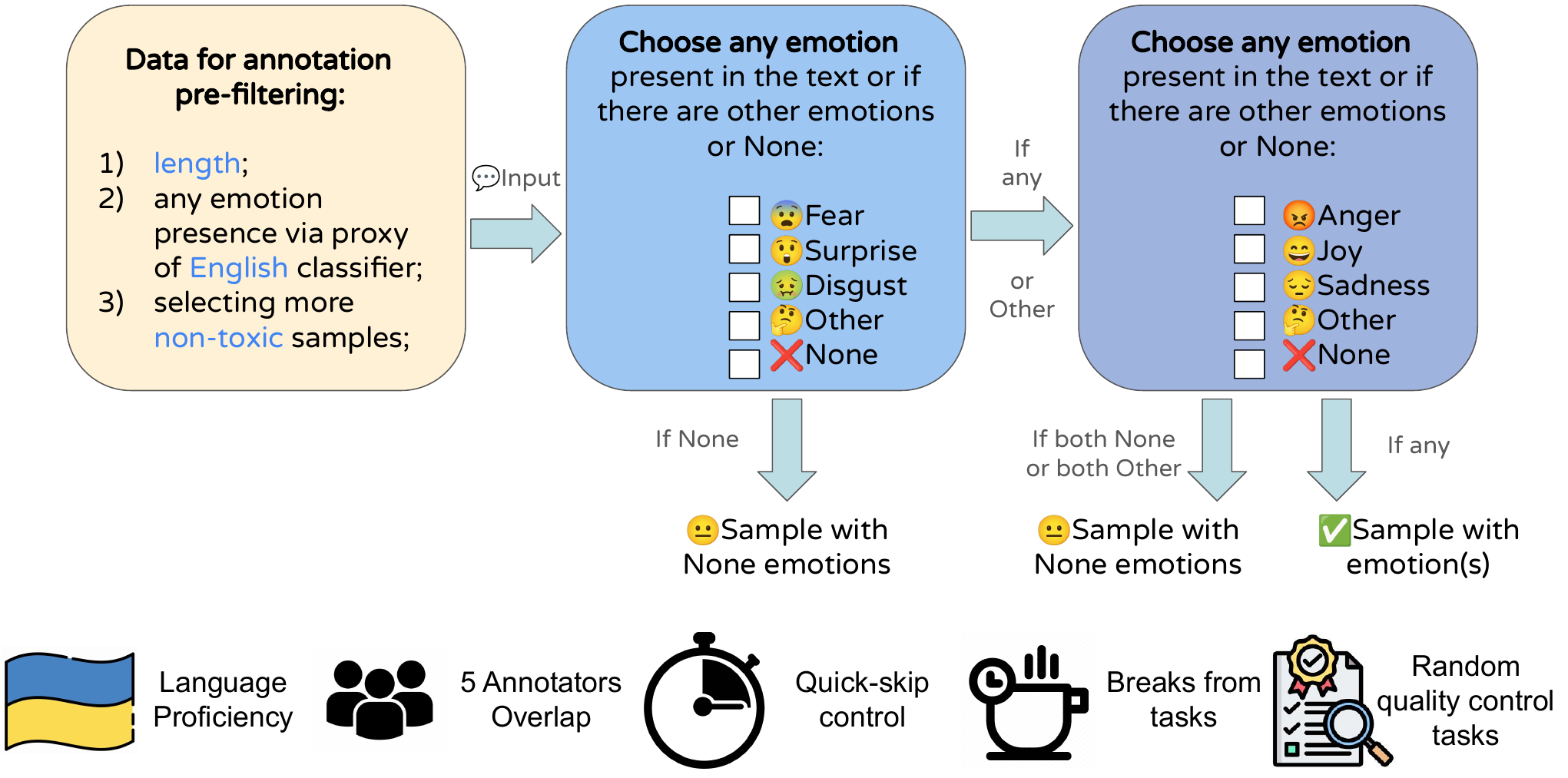}
    \caption{\emo Annotation Pipeline: we split the annotation into two tasks to improve annotator focus, and several quality control measures were applied to ensure the high quality of the collected data.}
    \label{fig:pipeline}
\end{figure*}

\begin{itemize}
    \item We design a robust \textbf{crowdsourcing annotation pipeline} for emotion annotation in Ukrainian texts, leveraging the Toloka.ai platform and incorporating quality control mechanisms to ensure high-quality annotations;
    \item Using this pipeline, we collect \textbf{EmoBench-UA}, the first manually annotated benchmark dataset for emotion detection in Ukrainian;\footnote{The dataset was part of SemEval-2025 shared Task 11~\cite{muhammad-etal-2025-semeval,muhammad-etal-2025-brighter} for the Ukrainian track.}
    \item We evaluate a comprehensive range of \textbf{classification approaches} on the dataset---including linguistic-based baselines, Transformer-based encoders, cross-lingual transfer methods, and prompting large language models (LLMs)---to assess task difficulty and provide a detailed performance analysis of the best to date emotion classifier in Ukrainian texts.
\end{itemize}

We release all the instructions, data, and top performing model fully online for public usage.\footnote{All resources with licenses are listed in Appendix~\ref{sec:our_resources}.} We believe that our insights into data annotation pipeline design for emotion detection, along with our experiments across different models, provide a reproducible foundation that can support the development of similar resources and technologies for other underrepresented languages where such corpora and emotion detection tools are not yet available.


\section{Related Work}

\paragraph{Emotion Detection Datasets and Models} As for many NLP tasks, various datasets, lexicon, and approaches in the first order were created for English emotion classification~\cite{mohammad-etal-2018-semeval}. Then, it was also extended to other popular languages like Spanish, German, and Arabic~\cite{plaza-del-arco-etal-2020-emoevent,chatterjee-etal-2019-semeval,DBLP:journals/kbs/KumarSA022} and then for some not so mainstream languages like Finish~\cite{ohman-etal-2020-xed}. Given the challenges associated with collecting fully annotated emotion datasets across languages, a multilingual emotional lexicon~\cite{mohammad-2023-best} which covers 100 languages was proposed by translating the original English resources, offering a practical first step toward facilitating emotion detection in lower-resource scenarios.

At the same time, the importance of developing robust NLP systems for emotion analysis and detection is well recognized~\cite{DBLP:journals/air/KusalPCKVP23}, especially in socially impactful domains such as customer service, healthcare, and support for minority communities. However, extending emotion detection capabilities uniformly across multiple languages remains a persistent challenge~\cite{de-bruyne-2023-paradox}. For English and several other languages, a variety of classification methods have been explored, ranging from BiLSTM and BERT-based models~\cite{al2020emodet2,de-bruyne-etal-2022-language} to more advanced architectures such as XLM-RoBERTa~\cite{conneau-etal-2020-unsupervised}, E5~\cite{DBLP:journals/corr/abs-2402-05672}, and multilingual LLMs like BLOOMz~\cite{wang-etal-2024-knowledge}.

\paragraph{Ukrainian Texts Classification} Although the availability of training data for classification tasks in Ukrainian remains limited, the research community has made notable strides in many NLP tasks. For example, UberText 2.0~\cite{chaplynskyi-2023-introducing} provides resources for NER tasks, legal document classification, and a wide range of textual sources including news, Wikipedia, and fiction. In addition, the OPUS corpus~\cite{tiedemann-2012-parallel} offers parallel Ukrainian data for cross-lingual applications. Recently, the Spivavtor dataset~\cite{saini-etal-2024-spivavtor} has also been introduced to facilitate instruction-tuning of Ukrainian-focused large language models.

For related classification tasks, resources for sentiment analysis~\cite{DBLP:conf/colins/ZalutskaMSMPBK23} have already been developed for Ukrainian. This task has received additional attention with more recently released dataset \texttt{COSMUS}~\cite{shynkarov-etal-2025-improving} with Ukrainian, Russian, and code-switch texts that reflect the real-life Ukrainian social networks communication diversity. The dataset cover four labels---\texttt{positive}, \texttt{negative}, \texttt{neutral}, and \texttt{mixed}---which provides a solid base for the first emotional level analysis.

From other styles perspective, toxicity classification corpus was introduced by~\citet{dementieva-etal-2024-toxicity}. Additionally, in the domain of abusive language, a bullying detection system for Ukrainian was developed but based on translated English data~\cite{oliinyk2023low}. Finally, \citet{dementieva-etal-2025-cross} explored various cross-lingual knowledge transfer methods for Ukrainian texts classification, yet emphasized the continued importance of authentic, manually annotated Ukrainian data. 
However, still, none of the released previously resources explicitly covered basic emotion detection task for the Ukrainian language.

\section{\emo Collection}
Here, we present the design of the crowdsourcing collection pipeline, detailing the task setup, annotation guidelines, interface design, and the applied quality control procedures used to obtain \emo. The overall schema of the pipeline is presented in Figure~\ref{fig:pipeline}.

\subsection{Emotion Classification Objective}
In this work, we define emotion recognition as the task of identifying perceived emotions---that is, the emotion that the majority of people would attribute to the speaker based on a given sentence or short text snippet~\cite{muhammad-etal-2025-brighter}.

We adopt the set of basic emotions proposed by~\citet{ekman1999basic}, which includes \texttt{Joy}, \texttt{Fear}, \texttt{Anger}, \texttt{Sadness}, \texttt{Disgust}, and \texttt{Surprise}. A single text instance may convey multiple emotions simultaneously creating the \textbf{multi-label} classification task. If a text \textit{does not express} any of the listed emotions, then we assign it the label \texttt{None}.

\subsection{Data Selection for Annotation}

\begin{figure}[ht!]
    \centering
    \includegraphics[width=0.8\linewidth]{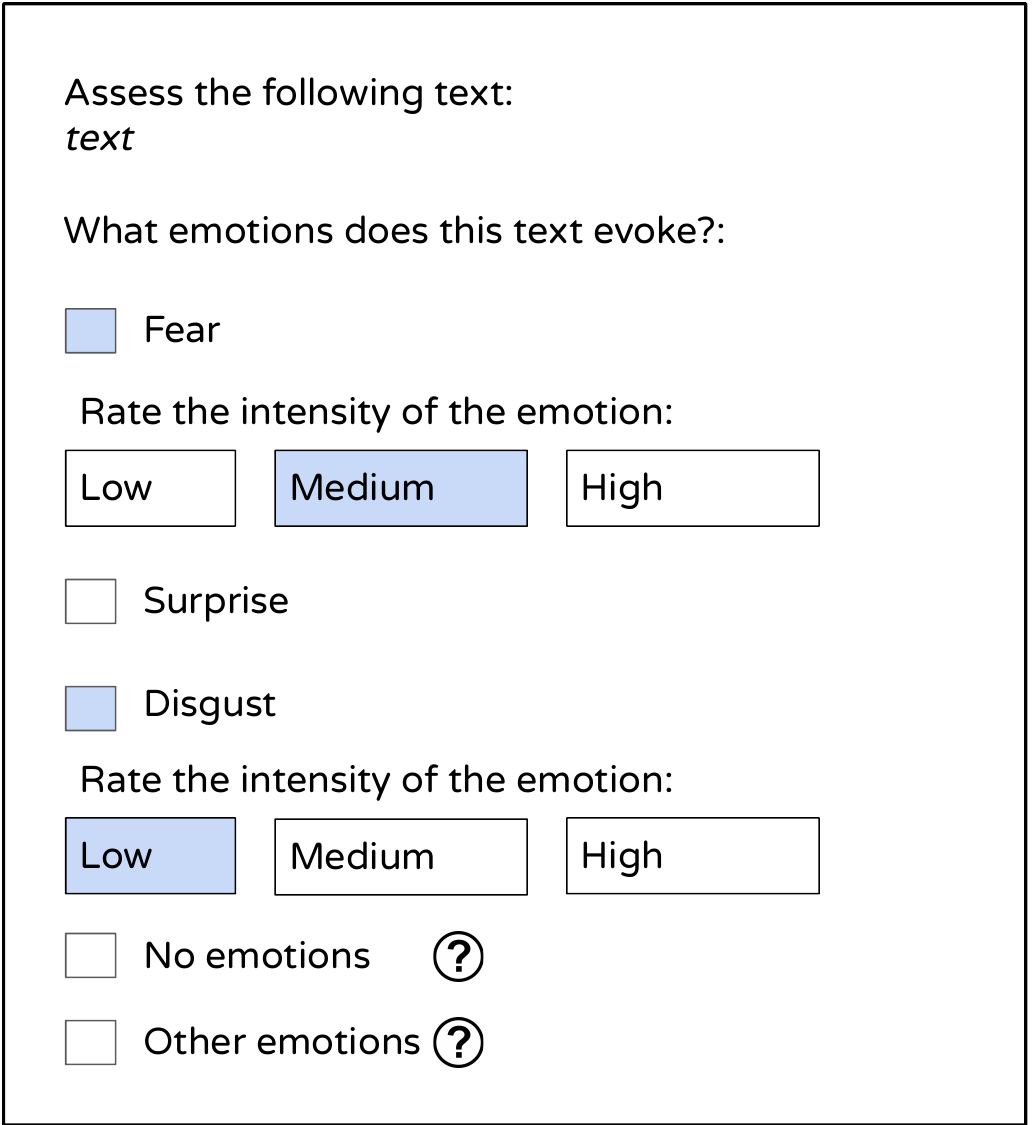}
    \caption{Annotation Interface illustration translated into English.}
    \label{fig:interface_eng}
\end{figure}

As the source data, we selected the publicly available Ukrainian tweets corpus~\cite{bobrovnik2019twt}. Given that social media posts are often rich in emotionally charged content, this corpus serves as a suitable foundation for our annotation task. Since the original collection consists of several hundred thousand tweets, we applied a multi-stage filtering process to both increase the likelihood of emotional content and ensure the feasibility of accurate annotation:
    
\paragraph{Length} First, we applied a length-based filter, discarding texts that were too short ($N$ words $<$ 5), as such samples often consist of hashtags or other non-informative tokens. Similarly, overly long texts ($N$ words $\geq$ 50) were excluded, as longer sequences tend to obscure the central meaning and make it challenging to accurately identify the expressed emotions.

\paragraph{Toxicity} While toxic texts can carry quite strong emotions, to ensure annotators well-being and general appropriateness of our corpus, we filtered out too toxic instances using opensourced toxicity classifier~\cite{dementieva-etal-2024-toxicity}.\footnote{\scriptsize{\href{https://huggingface.co/ukr-detect/ukr-toxicity-classifier}{https://huggingface.co/ukr-detect/ukr-toxicity-classifier}}}

\paragraph{Emotional Texts Pre-selection} To avoid an excessive imbalance toward emotionless texts, we performed a pre-selection step aimed at identifying texts likely to express emotions. Specifically, we applied the English emotion classifier \texttt{DistillRoBERTa-Emo-EN}\footnote{\scriptsize{\href{https://huggingface.co/michellejieli/emotion_text_classifier}{https://huggingface.co/michellejieli/emotion\_text\_classifier}}} on translated Ukrainian texts. For this, $10k$ Ukrainian samples, previously filtered by the previous steps, were translated into English using the NLLB model~\cite{costa2022no}\footnote{\scriptsize{\href{https://huggingface.co/facebook/nllb-200-distilled-600M}{https://huggingface.co/facebook/nllb-200-distilled-600M}}}. The emotion predictions from this classifier were then used to select a final set of $5k$ potentially emotionally-relevant texts, which were used for further annotation.


\subsection{Annotation Setup}
As emotion classification is quite subjective, we opted to rely on crowdsourcing rather than limiting the annotation process to a small group of expert annotators. For this, we utilized Toloka.ai\footnote{\scriptsize{\href{https://toloka.ai}{https://toloka.ai}}} crowdsourcing platform.

\subsubsection{Projects Design}
As shown in Figure~\ref{fig:pipeline}, to reduce cognitive load, we split the annotation process into two separate projects: one focused on \textit{fear}, \textit{surprise}, and \textit{disgust}; the other on \textit{anger}, \textit{joy}, and \textit{sadness}. Since our objective was to allow multiple emotion labels per instance, the projects were designed sequentially. 

Specifically, if a sample in the first stage received a label other than \texttt{None} (i.e., it was identified as containing emotion), it was subsequently included in the second stage. The final label assignment followed several scenarios: (i) samples could receive labels from both projects; (ii) samples could receive labels in the first project but none in the second, in which case only the first set of labels was retained; (iii) samples could receive labels other than our target emotions in the first project but valid labels in the second, in which case only the second set of labels was retained; and (iv) samples could be labeled as \texttt{Other} or \texttt{None} in both projects, in which case they were considered as containing no relevant emotions within our target categories.


\subsubsection{Instructions \& Interface}
Before being granted access to the annotation task, potential annotators were provided with detailed instructions, including a description of our aimed emotion detection task and illustrative examples for each emotion. We present the English translation of the introductory part of our instruction text:

\begin{tcolorbox}[colback=lightblue, colframe=blue!50, width=\linewidth, boxrule=0.5pt, arc=4pt, left=6pt, right=6pt, top=6pt, bottom=6pt]
\footnotesize{
\textbf{Instructions}\\
Select one or more emotions and their intensity in the text. If there are no emotions in the text or if there are emotions not represented in the list, select the No emotions/other emotions option.
}
\end{tcolorbox}

The full listed Ukrainian instructions for both projects are in Appendix~\ref{sec:app_instructions}. The English translation of the interface is presented in Figure~\ref{fig:interface_eng} with the original Ukrainian interface in Figure~\ref{fig:interface_uk}.

Annotators were instructed to answer a multiple-choice question, allowing them to select one or more emotions for each text instance. Additionally, they were asked to indicate the perceived intensity of the selected emotions. These annotations were also collected and will be included in the final release of \emo. However, for the purposes of this study in the experiments, we focus exclusively on the binary emotion presence labels.

\subsubsection{Annotators Selection}

\paragraph{Language Proficiency} Toloka platform provided pre-filtering mechanisms to select annotators who had passed official language proficiency tests, serving as an initial screening step. In our scenario, we selected annotators that were proficient in Ukrainian.

\paragraph{Training and Exam Phases} Annotators interested in participating first completed an unpaid training phase, where they reviewed detailed instructions and examples with explanations for correct labelling decisions. Following this, annotators were required to pass an exam, identical in format to the actual labelling tasks, to demonstrate their understanding of the guidelines. Successful candidates gained access to the main assignments.

\begin{figure*}[ht!]
    \hspace{-1.5cm}
    \begin{subfigure}[b]{0.49\textwidth}
        \centering
        \includegraphics[width=0.75\columnwidth]{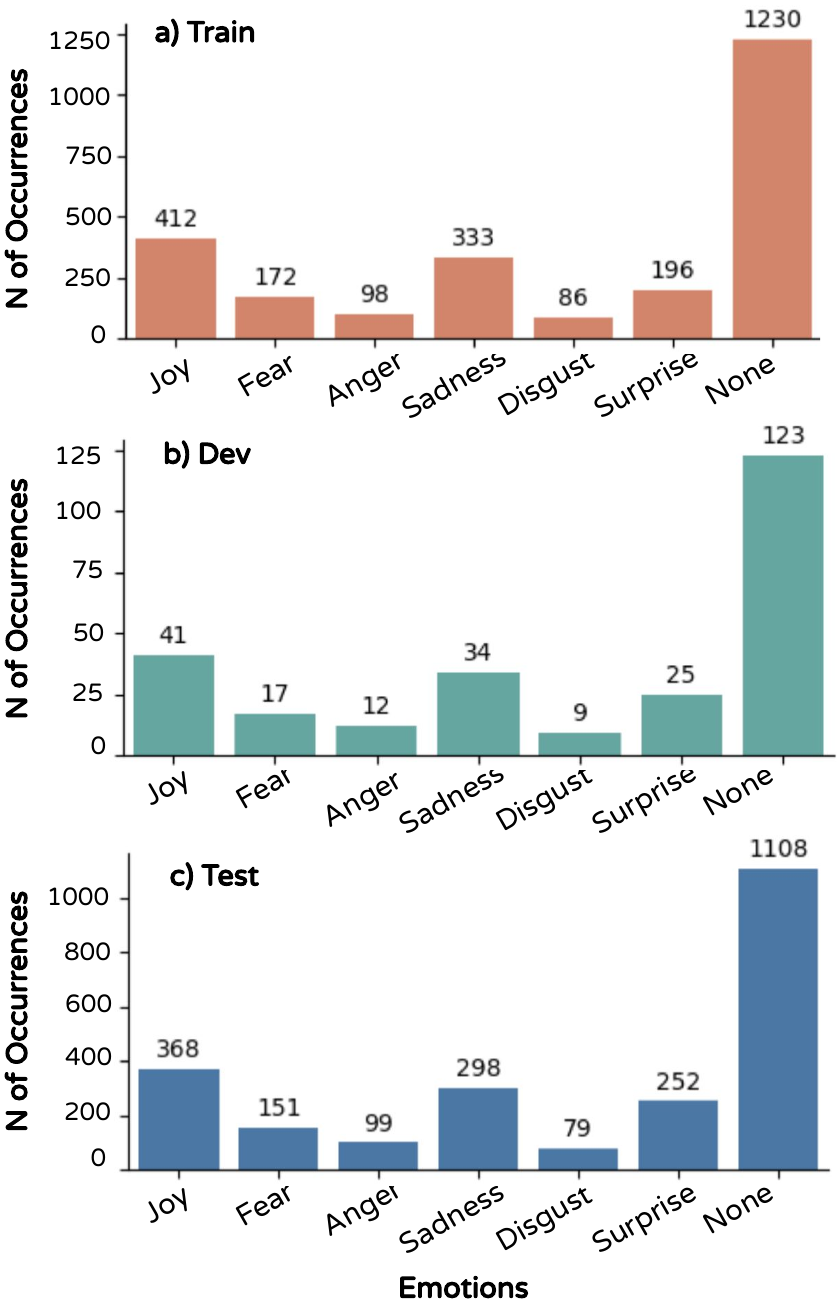}
        \caption{Distribution of Labels}
        \label{fig:labels_distribution}
    \end{subfigure}
    \hfill
    \begin{subfigure}[b]{0.49\textwidth}
        \hspace{-2cm}
        \includegraphics[width=1.3\columnwidth]{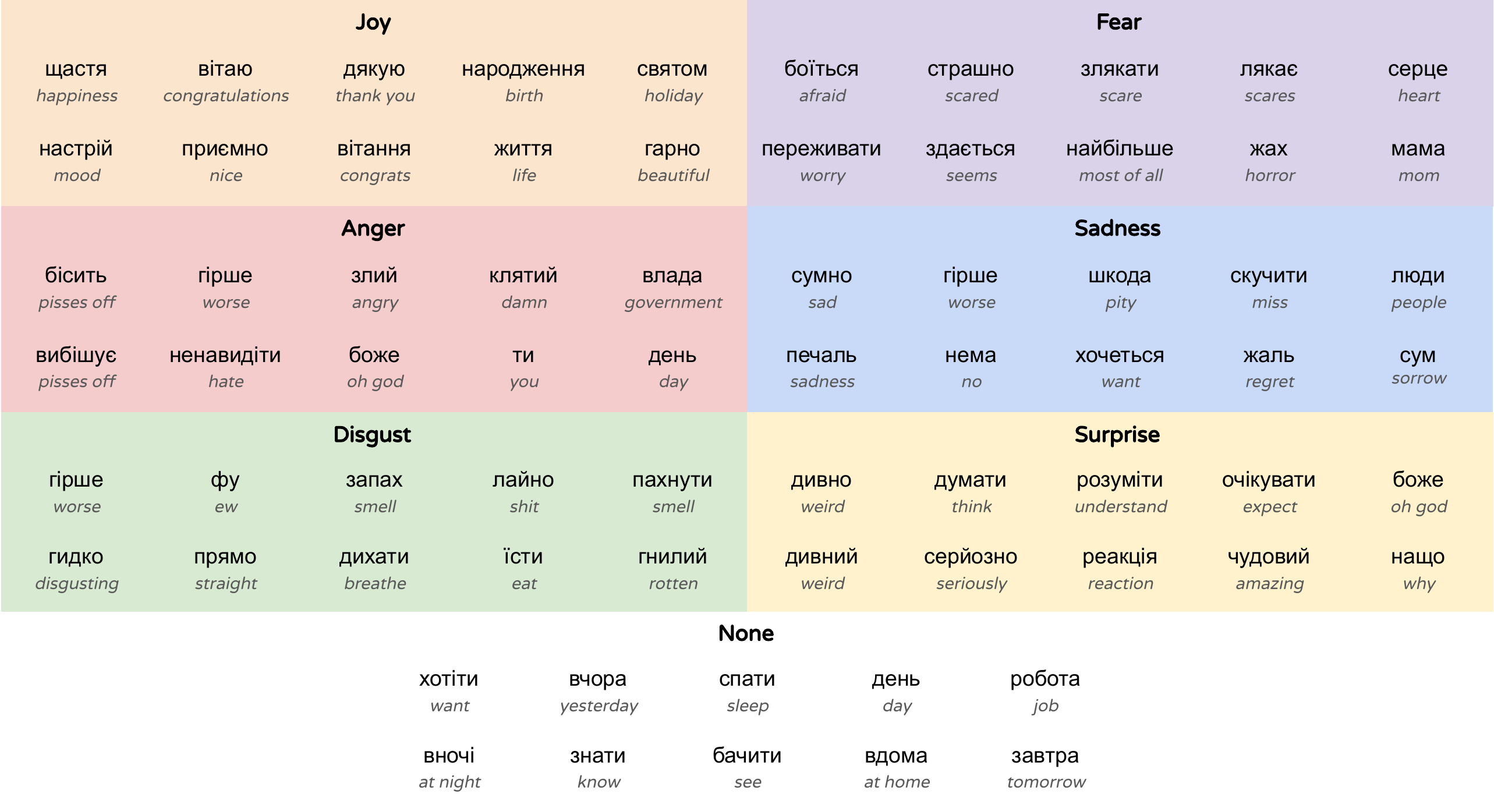}
        \vspace{1cm}
        \caption{Keywords per Emotion}
        \label{fig:keywords}
    \end{subfigure}

    \caption{\emo statistics per sets and emotions.}
\end{figure*}

\subsubsection{Quality Control}

To ensure high-quality annotations, we implemented several automated checks. Annotators were permanently banned if they submitted the last three task pages in under 15 seconds each, indicating low engagement. A one-day ban was triggered if three consecutive pages were skipped. To prevent fatigue, annotators were asked to take a 30-minute break after completing 25 consecutive pages. Additionally, control tasks were randomly injected; if the accuracy on these within the last 10 pages fell below 40\%, the annotator was temporarily banned and required to retake the training.

To ensure the reliability of the annotations, each text instance was labeled independently by \textbf{5 annotators}. The final emotion labels were determined through majority voting with an estimated confidence score by Dawid-Skene model~\cite{dawid1979maximum}. This aggregation model accounts not only for the annotators’ individual responses but also for their performance on control tasks, thereby weighting labels by annotator quality and improving overall robustness. Only instances with a confidence score $\geq$ 90\% from both projects were included to the final dataset.

\subsubsection{Annotators Well-Being}
\label{sec:annotators_wellbeing}

We aimed to design a fair, transparent, and user-friendly crowdsourcing project.

\paragraph{Fair Compensation} Payment rates were set to balance grant funding constraints with fair wages, aligning with Ukraine’s minimum hourly wage at the time of labelling (\textbf{1.12 USD/hour}). Annotators received \textbf{0.05 USD} per page with possibility to complete at least 20 assignment per hour. The overall spending of the whole project resulted in \textbf{500 USD}.

\paragraph{Positive Project Ratings} Toloka provided annotators with tools\footnote{\scriptsize{\href{https://toloka.ai/docs/guide/project_rating_stat/}{https://toloka.ai/docs/guide/project\_rating\_stat}}} to rate project fairness in terms of payment, task design, and organizer responsiveness. Our projects received high ratings: \textbf{4.80/5.00} for the Training Project and \textbf{4.90/5.00} for the Main Project.


\section{\emo}
After filtering out low-confidence and ambiguous samples from the annotation results, we obtained a final \emo of $4949$ labelled instances ($145$ samples were dropped due to label conflicts). Krippendorff's alpha agreement score was $0.85$. 
Then, we partitioned the dataset into fixed train/development/test subsets following a 50/5/45\% split ratio. An overview of the label distribution across these subsets is presented in Figure~\ref{fig:labels_distribution}. We also provide distribution of texts with one or more emotions per split in Appendix~\ref{sec:app_labels_cooccurrence}. The dataset examples can be found in Appendix~\ref{sec:app_samples}.

We were able to collect at least one hundred, and in some cases several hundred, instances for each emotion category. The majority of the sentences have clear distinguished one emotion with rare cases of texts with several emotions. Nevertheless, the dataset remains imbalanced, with \texttt{Joy} and \texttt{Sadness} being the most prevalent emotions among the labeled samples, alongside a substantial portion of texts assigned the \texttt{None} label. Such imbalance is a common characteristic of emotion detection datasets, reflecting the natural distribution of emotions in real-world text and contributing to the overall complexity of the task.

Additionally, in Figure~\ref{fig:keywords}, we provide a closer analysis of the collected emotional data by extracting the top-10 keywords for each emotion label (lemmatization done using the \texttt{spacy}\footnote{\scriptsize{\href{https://spacy.io/models/uk}{https://spacy.io/models/uk}}} library). The resulting keywords reveal clear and intuitive associations with the corresponding emotional categories, further confirming the quality and relevance of the annotated texts. 

\section{Models}
We test various types on models on our collected dataset: (i)~linguistic-based approaches; (ii)~Transformer-based encoders; (iii)~LLMs prompting for classification. Then, we also did an ablation study with synthetic Ukrainian training data acquisition via translation from English. The details of models' hyperparameters can be found in Appendix~\ref{sec:app_hyperparameters}.

\subsection{Linguistic-based Approaches}
Even with current advances in NLP, linguistic-based approaches based on statistics of the training set can be quite a strong and resource-efficient baseline for stylistic texts classification like sentiment~\cite{DBLP:journals/csur/BrauwersF23} or formality~\cite{dementieva-etal-2023-detecting}.

\paragraph{Keywords Based}  We used the train part of our dataset to extract \textit{natural} keywords per emotion as shown in Figure~\ref{fig:keywords}. We used \texttt{spacy} for lemmatization extracting top-20 words per emotion. For each text, we count the number of keywords associated with each emotion and assign the emotion with the highest keyword frequency.

\paragraph{Logistic Regression} Firstly, we embed our texts with \texttt{CountVectorizer} into \texttt{td-idf} features. Then, we fine-tuned Logistic Regressions classifier on the training part of our dataset.

\paragraph{Random Forest} The same as for logistic regression, we fine-tune Random Forest classifier with $100$ decision trees on \texttt{td-idf} training features.

\subsection{Transformer-based Encoder}
Then, we take the next generation of classification models based on the Transformers~\cite{DBLP:conf/nips/VaswaniSPUJGKP17} encoders. For each model type, we evaluate multiple versions varying in model size.

\paragraph{BERT} Firstly, we used \texttt{mBERT}\footnote{\scriptsize{\href{https://huggingface.co/google-bert/bert-base-multilingual-cased}{https://huggingface.co/google-bert/bert-base-multilingual-cased}}}~\cite{DBLP:journals/corr/abs-1810-04805} as it contains Ukrainian in the pre-trained data. We additionally experimented with a compact variant---\texttt{Geotrend-BERT}\footnote{\scriptsize{\href{https://huggingface.co/Geotrend/bert-base-uk-cased}{https://huggingface.co/Geotrend/bert-base-uk-cased}}}---of mBERT where the vocabulary and embeddings were specifically refined to retain only Ukrainian~\cite{abdaoui-etal-2020-load}.

\paragraph{RoBERTa} As an extension of BERT-alike models, we used several versions of RoBERTa-alike models~\cite{DBLP:journals/corr/abs-1911-02116} as it shown previously promising results in Ukrainian texts classification~\cite{dementieva-etal-2025-cross}:
\begin{itemize}[leftmargin=10pt]
\itemsep-0.2em  
    \item \texttt{XLM-RoBERTa}: \texttt{base}\footnote{\scriptsize{\href{https://huggingface.co/FacebookAI/xlm-roberta-base}{https://huggingface.co/FacebookAI/xlm-roberta-base}}} and \texttt{large}\footnote{\scriptsize{\href{https://huggingface.co/FacebookAI/xlm-roberta-large}{https://huggingface.co/FacebookAI/xlm-roberta-large}}} instances;
    \item Ukrainian-specific pre-trained monolingual RoBERTa: \texttt{UKR-RoBERTa-base}\footnote{\scriptsize{\href{https://huggingface.co/youscan/ukr-roberta-base}{https://huggingface.co/youscan/ukr-roberta-base}}};
    \item additionally fine-tuned on sentiment classification task on Twitter data \texttt{Twitter-XLM-RoBERTa base}\footnote{\scriptsize{\href{https://huggingface.co/cardiffnlp/twitter-xlm-roberta-base-sentiment}{https://huggingface.co/cardiffnlp/twitter-xlm-roberta-base-sentiment}}}~\cite{barbieri-etal-2022-xlm};
    \item finally, we tested \texttt{Glot500-base}\footnote{\scriptsize{\href{https://huggingface.co/cis-lmu/glot500-base}{https://huggingface.co/cis-lmu/glot500-base}}} model~\cite{imanigooghari-etal-2023-glot500} that extended multilingual RoBERTa to 500 languages.
\end{itemize}

\paragraph{LaBSe} Another multilingual embedding model covering 109 languages including Ukrainian: \texttt{LaBSe}\footnote{\scriptsize{\href{https://huggingface.co/sentence-transformers/LaBSE}{https://huggingface.co/sentence-transformers/LaBSE}}}~\cite{feng-etal-2022-language}.

\paragraph{E5} Finally, we utilized the more recent \texttt{multilingual-e5} embeddings~\cite{DBLP:journals/corr/abs-2402-05672}: \texttt{base}\footnote{\scriptsize{\href{https://huggingface.co/intfloat/multilingual-e5-base}{https://huggingface.co/intfloat/multilingual-e5-base}}} and \texttt{large}\footnote{\scriptsize{\href{https://huggingface.co/intfloat/multilingual-e5-large}{https://huggingface.co/intfloat/multilingual-e5-large}}} variants.

\subsection{LLMs prompting}
To test models based on another methodology, we also tried out various modern LLMs on our benchmark dataset transforming our classification task into the \texttt{text-to-text} generation one. While Ukrainian is not always explicitly present in the pre-training data reports, the emerging abilities of LLMs already showed promising results in handling new languages~\cite{DBLP:journals/tmlr/WeiTBRZBYBZMCHVLDF22} including Ukrainian~\cite{dementieva-etal-2025-cross}. However, we also utilize more recent LLMs dedicated to European languages, including Ukrainian. We used two types of prompts---instructions in English and in Ukrainian---that are fully listed in Appendx~\ref{sec:app_prompts}.



We tested several families of LLMs with variants in terms of version and sizes. We chose mostly instruction tuned instances as they supposedly perform more precise for classification tasks:

\paragraph{EuroLLM} The recent initiative introduced in~\cite{DBLP:journals/corr/abs-2409-16235} has an aim to develop high-quality LLMs for European languages with Ukrainian definitely included. We selected \texttt{EuroLLM-1.7B}\footnote{\scriptsize{\href{https://huggingface.co/utter-project/EuroLLM-1.7B-Instruct}{https://huggingface.co/utter-project/EuroLLM-1.7B-Instruct}}} variant for our experiments.

\paragraph{Spivavtor} Ukrainian-tuned LLM that was obtained by instruction tuning \texttt{CohereForAI/aya-101}~\cite{DBLP:conf/acl/UstunAYKDOBSOKV24} model on the Spivavtor dataset~\cite{saini-etal-2024-spivavtor}.\footnote{\scriptsize{\href{https://huggingface.co/grammarly/spivavtor-xxl}{https://huggingface.co/grammarly/spivavtor-xxl}}}

\paragraph{MamayLM} Another specifically Ukrainian-tuned LLM obtained from \texttt{Gemma-2}~\cite{DBLP:journals/corr/abs-2408-00118} and, in 2025, achieved the top scores within Ukrainian LLMs. It was continuously pre-trained on a large pre-filtered dataset (75B tokens of Ukrainian and English data in total) using the combination of data mixing and model merging~\cite{MamayLM}.\footnote{\scriptsize{\href{https://huggingface.co/INSAIT-Institute/MamayLM-Gemma-2-9B-IT-v0.1}{https://huggingface.co/INSAIT-Institute/MamayLM-Gemma-2-9B-IT}}}

\paragraph{Mistral} From multilingual general-purpose LLMs, firstly, we used several version of Mistral-family models~\cite{Jiang2023Mistral7}---\texttt{Mistral-7B}\footnote{\scriptsize{\href{https://huggingface.co/mistralai/Mistral-7B-Instruct-v0.3}{https://huggingface.co/mistralai/Mistral-7B-Instruct-v0.3}}}
and \texttt{Mixtral-8x7B}.\footnote{\scriptsize{\href{https://huggingface.co/mistralai/Mixtral-8x7B-Instruct-v0.1}{https://huggingface.co/mistralai/Mixtral-8x7B-Instruct-v0.1}}} The models cards do not mention explicitly Ukrainian and other languages, however Mistral showed promising results in Ukrainian texts classification tasks~\cite{dementieva-etal-2025-cross}.

\paragraph{LLaMa3} The LLaMa model~\cite{llama3modelcard} card as well does not stated Ukrainian explicitly, however, encourages research in usage of the model in various multilingual tasks. Thus, we tested the   \texttt{Llama-3-8B}\footnote{\scriptsize{\href{https://huggingface.co/meta-llama/Meta-Llama-3-8B-Instruct}{https://huggingface.co/meta-llama/Meta-Llama-3-8B-Instruct}}} and \texttt{Llama-3.3-70B}\footnote{\scriptsize{\href{https://huggingface.co/meta-llama/Llama-3.3-70B-Instruct}{https://huggingface.co/meta-llama/Llama-3.3-70B-Instruct}}} variants.

\paragraph{Qwen3} Then, we also utilized for experiments one of the \texttt{Qwen3}~\cite{qwen3technicalreport} family of models\footnote{\scriptsize{\href{https://huggingface.co/Qwen/Qwen3-4B-Instruct-2507-FP8}{https://huggingface.co/Qwen/Qwen3-4B-Instruct-2507-FP8}}} that showed before promising results in various Ukrainian understanding tasks~\cite{kravchenko-etal-2025-ualign}. 

\paragraph{DeepSeek} Finally, we tested one of the recent top performing models in reasoning---DeepSeek~\cite{DBLP:journals/corr/abs-2501-12948} with \texttt{DeepSeek-R1-Qwen}\footnote{\scriptsize{\href{https://huggingface.co/deepseek-ai/DeepSeek-R1-Distill-Qwen-7B}{https://huggingface.co/deepseek-ai/DeepSeek-R1-Distill-Qwen-7B}}}, \texttt{deepseek-ai/DeepSeek-R1-Llama}\footnote{\scriptsize{\href{https://huggingface.co/deepseek-ai/DeepSeek-R1-Distill-Llama-8B}{https://huggingface.co/deepseek-ai/DeepSeek-R1-Distill-Llama-8B}}}, and \texttt{DeepSeek-V3}\footnote{\scriptsize{\href{https://huggingface.co/deepseek-ai/DeepSeek-V3}{https://huggingface.co/deepseek-ai/DeepSeek-V3}}} variants. The situation of the Ukrainian language presence in the models is the same as for Mistral and LLaMa---DeepSeek was heavily optimized for English and Chinese, however, the authors encourage to try it for other languages.

\subsubsection{Translation \& Synthetic Data}
Additionally, we also experimented with cross-lingual setups to imitate various low-resource scenarios: (i) translation in \texttt{ukr$\rightarrow$en} direction; (ii) translation in \texttt{en$\rightarrow$ukr} direction.

\begingroup
\renewcommand{\arraystretch}{1.15}
\begin{table*}[ht!]
    \centering
    \scriptsize
    \begin{tabular}{l|c c c c c c c | c c c}
    \toprule
     & \textbf{Joy} & \textbf{Fear} & \textbf{Anger} & \textbf{Sadness} & \textbf{Disgust} & \textbf{Surprise} & \textbf{None} & \textbf{Pr} & \textbf{Re} & \textbf{F1} \\
    \midrule
    \multicolumn{11}{c}{\textbf{\textit{Linguistic-based Approaches}}} \\
    \midrule
    Keywords & 0.30 & 0.15 & 0.08 & 0.21 & 0.10 & 0.15 & 0.25 & 0.24 & 0.24 & 0.22\\
    Logistic Regression & \textbf{0.64} & \textbf{0.72} & \textbf{0.49} & \textbf{0.59} & \textbf{0.49} & \cellcolor{LightOrange}\textbf{0.61} & 0.67 & 0.51 & \textbf{0.22} & \textbf{0.29} \\
    Random Forest & 0.61 & 0.69 & \textbf{0.49} & \textbf{0.59} & \textbf{0.49} & 0.60 & \textbf{0.68} & \textbf{0.58} & 0.21 & 0.27\\
    \midrule
    \multicolumn{11}{c}{\textbf{\textit{Translation to English}}} \\
    \midrule
    DistillRoBERTa-Emo-EN & 0.56 & 0.55 & 0.31 & 0.52 & 0.23 & 0.47 & 0.55 & 0.40 & 0.61 & 0.45 \\
    \midrule
    \multicolumn{11}{c}{\textbf{\textit{Transformer-based Encoders}}} \\
    \midrule
    LaBSe & 0.67 & 0.73 & 0.30 & 0.65 & 0.33 & 0.54 & 0.80 & 0.57 & 0.59 & 0.57 \\
    \hline
    Geotrend-BERT & \textbf{0.58} & \textbf{0.59} & \textbf{0.08} & \textbf{0.50} & \textbf{0.11} & \textbf{0.40} & \textbf{0.73} & \textbf{0.46} & \textbf{0.43} & \textbf{0.43} \\
    mBERT & 0.46 & 0.24 & 0.01 & 0.45 & 0.02 & 0.33 & 0.73 & 0.33 & 0.33 & 0.32\\
    \hline
    UKR-RoBERTa Base & 0.65 & 0.58 & 0.14 & 0.50 & \textbf{0.21} & 0.49 & 0.74 & 0.51 & 0.45 & 0.47 \\
    XLM-RoBERTa Base & 0.61 & 0.31 & 0.00 & 0.33 & 0.01 & 0.19 & 0.75 & 0.33 & 0.31 & 0.31 \\
    XLM-RoBERTa Large & \cellcolor{LightOrange}\textbf{0.73} & \textbf{0.79} & \textbf{0.20} & \textbf{0.68} & 0.00 & \textbf{0.60} & \textbf{0.80} & 0.52 & \textbf{0.58} & \textbf{0.54} \\
    Twitter-XLM-RoBERTa & 0.72 & 0.76 & 0.13 & 0.64 & 0.07 & 0.54 & 0.79 & \cellcolor{LightOrange}\textbf{0.66} & 0.51 & 0.52 \\
    Glot500 Base & 0.01	& 0.02 & 0.03 & 0.18 & 0.00 & 0.01 & 0.64 & 0.24 & 0.19 & 0.13 \\
    \hline
    Multilingual-E5 Base & 0.71 & 0.73 & 0.01 & 0.52 & 0.00 & 0.50 & 0.77 & 0.49 & 0.45 & 0.46 \\
    Multilingual-E5 Large & \cellcolor{LightOrange}\textbf{0.73} & \cellcolor{LightOrange}\textbf{0.81} & \textbf{0.31} & \textbf{0.69} & \textbf{0.35} & \textbf{0.60} & \cellcolor{LightOrange}\textbf{0.81} & \textbf{0.65} & \textbf{0.62} & \textbf{0.62} \\ 
    \midrule
    \multicolumn{11}{c}{\textbf{\textit{LLMs Prompting}}} \\
    \midrule
    EuroLLM-1.7B (ENG) & \textbf{0.46} & \textbf{0.31} & \textbf{0.15} & \textbf{0.37} & \textbf{0.18} & 0.09 & \textbf{0.28} & \textbf{0.26} & \textbf{0.38} & \textbf{0.26}\\
    EuroLLM-1.7B (UKR) & 0.38 & 0.30 & 0.11 & 0.27 & 0.10 & \textbf{0.11} & 0.25 & 0.25 & 0.24 & 0.22 \\
    \hline
    Spivavtor-XXL (ENG) & \textbf{0.39} & 0.03 & \textbf{0.15} & \textbf{0.13} & 0.00 & 0.01 & \textbf{0.69} & \textbf{0.68} & 0.20 & \textbf{0.20}\\
    Spivavtor-XXL (UKR) & 0.32 & \textbf{0.08} & 0.14 & \textbf{0.13} & \textbf{0.08} & \textbf{0.13} & 0.29 & 0.17 & \textbf{0.28} & 0.17 \\
    \hline
    MamayLM-9B (ENG) & \textbf{0.63} & \textbf{0.62} & \textbf{0.54} & \textbf{0.64} & 0.38 & \textbf{0.31} & \textbf{0.67} & \textbf{0.46} & \textbf{0.73} & \textbf{0.54} \\
    MamayLM-9B (UKR) & 0.61 & 0.61 & 0.47 & 0.52 & \textbf{0.47} & 0.24 & 0.41 & 0.44 & \textbf{0.73} & 0.48 \\
    \hline
    Mistral-7B (ENG) & 0.52 & \textbf{0.58} & \textbf{0.33} & \textbf{0.49} & \textbf{0.32} & \textbf{0.37} & \textbf{0.52} & \textbf{0.37} & \textbf{0.73} & \textbf{0.45} \\
    Mistral-7B (UKR) & \textbf{0.55} & 0.37 & 0.28 & 0.47 & 0.19 & 0.24 & 0.33 & 0.32 & 0.71 & 0.35\\
    \hline
    Mixtral-8x7B (ENG) & \textbf{0.49} & \textbf{0.37} & \textbf{0.34} & \textbf{0.51} & \textbf{0.25} & \textbf{0.25} & 0.66 & \textbf{0.32} & \textbf{0.74} & \textbf{0.41} \\
    Mixtral-8x7B (UKR) & 0.48 & 0.35 & 0.19 & 0.47 & 0.21 & 0.22 & \textbf{0.71} & 0.27 & 0.73 & 0.37 \\
    \hline
    LLaMA 3 8B (ENG) & \textbf{0.56} & 0.65 & \textbf{0.36} & \textbf{0.54} & \textbf{0.29} & \textbf{0.25} & \textbf{0.39} & \textbf{0.43} & \textbf{0.56} & \textbf{0.43}\\
    LLaMA 3 8B (UKR) & 0.30 & \textbf{0.67} & 0.29 & 0.45 & 0.15 & \textbf{0.25} & 0.10 & 0.38 & 0.53 & 0.31 \\
    \hline
    LLaMA 3.3 70B (ENG) & \textbf{0.64} & 0.63 & \textbf{0.47} & 0.62 & \textbf{0.26} & 0.32 & \textbf{0.43} & 0.44 & \textbf{0.79} & \textbf{0.48} \\
    LLaMA 3.3 70B (UKR) & 0.58 & \textbf{0.68} & 0.34 & \textbf{0.71} & 0.18 & \textbf{0.33} & 0.36 & \textbf{0.45} & 0.64 & 0.46 \\
    \hline
    Qwen3-4B (ENG) & \textbf{0.65} & \textbf{0.66} & 0.39 & \textbf{0.56} & \textbf{0.34} & \textbf{0.35} & \textbf{0.52} & \textbf{0.45} & \textbf{0.72} & \textbf{0.49} \\
    Qwen3-4B (UKR) & 0.63 & 0.62 & \textbf{0.42} & 0.54 & 0.18 & 0.34 & 0.33 & 0.43 & 0.69 & 0.44 \\
    \hline
    DeepSeek-R1-Qwen (ENG) & 0.63 & 0.61 & \textbf{0.43} & \textbf{0.64} & \textbf{0.45} & \textbf{0.46} & 0.60 & \textbf{0.48} & \textbf{0.75} & \textbf{0.55} \\
    DeepSeek-R1-Qwen (UKR) & \textbf{0.68} & \textbf{0.66} & 0.40 & 0.57 & 0.29 & 0.38 & \textbf{0.68} & 0.46 & 0.66 & 0.52 \\
    \hline
    DeepSeek-R1-LLaMA (ENG) & \textbf{0.67} & \textbf{0.69} & \textbf{0.49} & \textbf{0.71} & \textbf{0.52} & 0.47 & 0.67 & \textbf{0.54} & \textbf{0.72} & \textbf{0.60} \\
    DeepSeek-R1-LLaMA (UKR) & \textbf{0.67} & 0.64 & 0.45 & 0.69 & 0.33 & \textbf{0.51} & \textbf{0.69} & 0.51 & 0.69 & 0.57 \\
    \hline
    DeepSeek-V3 (ENG) & \cellcolor{LightOrange}\textbf{0.73} & \textbf{0.74} & 0.60 & \cellcolor{LightOrange}\textbf{0.72} & \cellcolor{LightOrange}\textbf{0.57} & 0.41 & \textbf{0.78} & \textbf{0.60} & 0.72 & \cellcolor{LightOrange}\textbf{0.65} \\
    DeepSeek-V3 (UKR) & 0.71 & 0.66 & \cellcolor{LightOrange}\textbf{0.61} & \cellcolor{LightOrange}\textbf{0.72} & 0.48 & \textbf{0.42} & 0.71 & 0.54 & \cellcolor{LightOrange}\textbf{0.81} & 0.62 \\
    \bottomrule
    \end{tabular}
    \caption{\emo test set results of various models types per emotion and overall. The models that required fine-tuning were trained on the natural \textbf{Ukrainian} training set of \emo. Per emotion, we report F1 scores. In \textbf{bold}, we denote the best results per column within model type. In \colorbox{LightOrange}{orange} we highlight the top results per column.}
    \label{tab:results_final}
\end{table*}
\endgroup

\paragraph{Synthetic Emotion Lexicon} In addition to natural Ukrainian lexicon extracted from our data, we also experimented with the already collected and translated from English \textit{synthetic} Ukrainian emotions lexicon from~\cite{mohammad-2023-best}.

\paragraph{Translation} Then, we imitated the scenario if we have already fine-tuned English emotion detection model---i.e. \texttt{DistillRoBERTa-Emo-EN}\footnote{\scriptsize{\href{https://huggingface.co/michellejieli/emotion_text_classifier}{https://huggingface.co/michellejieli/emotion\_text\_classifier}}}---so then we can translate Ukrainian inputs into English to obtain the labels.

\paragraph{Synthetic Training Data via Translation} To not rely on the translation at every single inference, we can also translate the whole English training corpus~\cite{muhammad-etal-2025-semeval} into Ukrainian and then used it as Ukrainian training data.

For translation in all scenarios, we utilized NLLB\footnote{\scriptsize{\href{https://huggingface.co/facebook/nllb-200-distilled-600M}{https://huggingface.co/facebook/nllb-200-distilled-600M}}} model~\cite{costa2022no}.

\section{Results}

\begingroup
\renewcommand{\arraystretch}{1.15}
\begin{table*}[ht!]
    \centering
    \scriptsize
    \begin{tabular}{l|c c c c c c | c c c}
    \toprule
     & \textbf{Joy} & \textbf{Fear} & \textbf{Anger} & \textbf{Sadness} & \textbf{Surprise} & \textbf{None} & \textbf{Pr} & \textbf{Re} & \textbf{F1} \\
    \midrule
    Keywords UK & \textbf{0.30} & \textbf{0.15} & \textbf{0.08} & \textbf{0.21} & \textbf{0.15} & \textbf{0.25} & \textbf{0.27} & \textbf{0.25} & \textbf{0.26}\\
    Keywords EN & 0.17 & 0.05 & 0.01 & 0.18 & 0.08 & 0.11 & 0.15 & 0.01 & 0.10\\
    \midrule
    UKR-RoBERTa-base UK & \textbf{0.65} & \textbf{0.58} & 0.14 & \textbf{0.50} & \textbf{0.49} & \textbf{0.74} & \textbf{0.56} & \textbf{0.49} & \textbf{0.52} \\
    UKR-RoBERTa-base EN & 0.53 & 0.24 & \textbf{0.19} & 0.30 & 0.31 & 0.60 & 0.32 & 0.42 & 0.36 \\
    \midrule
    mBERT UK & \textbf{0.46} & \textbf{0.24} & 0.00 & \textbf{0.45} & 0.33& \textbf{0.73} & \textbf{0.38} & \textbf{0.38} & \textbf{0.37} \\
    mBERT EN & 0.38 & 0.12 & \textbf{0.12} & 0.31 & \textbf{0.31} & 0.55 & 0.31 & 0.30 & 0.30 \\
    \midrule
    LaBSe UK & \textbf{0.67} & \textbf{0.73} & \textbf{0.30} & \textbf{0.65} & \textbf{0.54} & \textbf{0.80} & \textbf{0.59} & \textbf{0.65} & \textbf{0.62}\\
    LaBSE EN & 0.60 & 0.41 & 0.22 & 0.39 & 0.30 & 0.64 & 0.44 & 0.43 & 0.43\\
    \midrule
    XLM-RoBERTa Large UK & \textbf{0.73} & \textbf{0.79} & \textbf{0.20} & \textbf{0.68} & \textbf{0.60} & \textbf{0.80} & \textbf{0.61} & \textbf{0.68} & \textbf{0.63}\\
    XLM-RoBERTa Large EN & 0.50 & 0.34 & 0.15 & 0.47 & 0.24 & 0.53 & 0.33 & 0.45 & 0.37\\
    \midrule
    Twitter-XLM-RoBERTa UK & \textbf{0.72} & \textbf{0.76} & 0.13 & \textbf{0.64} & \textbf{0.54} & \textbf{0.79} & \textbf{0.60} & \textbf{0.59} &  \textbf{0.60}\\
    Twitter-XLM-RoBERTa EN & 0.62 & 0.26 & \textbf{0.21} & 0.52 & 0.44 & 0.62 & 0.42 & 0.47 & 0.44\\
    \midrule
    Multilingual-E5 Large UK & \textbf{0.73} & \textbf{0.81} & \textbf{0.31} & \textbf{0.69} & \textbf{0.60} & \textbf{0.81} & \textbf{0.65} & \textbf{0.68} & \textbf{0.66}\\ 
    Multilingual-E5 Large EN & 0.61 & 0.26 & 0.22 & 0.36 & 0.23 & 0.56 & 0.36 & 0.41 & 0.37\\ 
    \bottomrule
    \end{tabular}
    \caption{\emo test set results of comparison natural \texttt{UK} vs synthetic translated from \texttt{EN} training data. Per emotion, we report F1 scores. In \textbf{bold}, we denote the best results per column within model type. As the English dataset does not contain \texttt{Disgust} label, we fine-tuned all models types without it for this experiment.}
    \label{tab:results_synthetic}
\end{table*}
\endgroup

The results of models evaluation on the test part of our novel \emo dataset on the \textbf{binary multi-label classification} task are presented in the Table~\ref{tab:results_final}. We report \textbf{F1 score} per each emotion; for overall results, we report Precision, Recall, and \textbf{macro-averaged F1-score}. Also, we provide the confusion matrices for the top performing models in Appendix~\ref{sec:app_conf_matrices}.

\paragraph{Linguistic-based Approaches} While the linguistic-based models rely on relatively simple statistical representations of the text, they demonstrate competitive performance. The keyword-based approach, however, yielded lower results, which is expected given that emotion detection often relies on understanding contextual collocations and multi-word expressions rather than isolated words. In contrast, both logistic regression and random forest models performed on par with several \texttt{base} encoder models and, in some cases, even outperformed certain LLMs. Although these models did not achieve the highest overall F1-macro scores, they showed strong precision but struggled with recall.

\paragraph{Transformer-based Encoders} Among the range of tested BERT- and RoBERTa-based models, the Ukrainian-specific encoders, \texttt{Geotrend-BERT} and \texttt{UKR-RoBERTa Base}, significantly outperformed \texttt{mBERT}, \texttt{Glot500}, and \texttt{XLM-RoBERTa-base}, highlighting the importance of monolingual, Ukrainian-specific encoders. At the same time, the multilingual \texttt{LaBSE} model outperformed Ukrainian-specific models. Within the RoBERTa-like family, \texttt{XLM-RoBERTa-large} and \texttt{Twitter-XLM-RoBERTa} achieved the strongest results, although both struggled with the \texttt{Anger} and \texttt{Disgust}. Finally, the best-performing encoder was \textbf{\texttt{Multilingual-E5-Large}}, with a good balance of Precision and Recall.

\paragraph{LLMs} Across all model families, we observe a consistent trend of slightly improved performance when models are prompted in English rather than Ukrainian. Surprisingly, \texttt{EuroLLM} underperformed, yielding results even lower than the linguistics-based baselines. Other LLMs delivered scores comparable to encoder-based models, outperforming them in the \texttt{Anger} and \texttt{Disgust} classes. While all LLMs demonstrated lower Precision compared to the best encoders, they consistently achieved higher Recall. Notably, \textbf{\texttt{DeepSeek-V3}} handled the emotion detection task in Ukrainian with the highest scores. However, the overall performance gains over \texttt{Multilingual-E5-Large} remain minimal, raising a question regarding the efficiency and responsible usage of such large models.

\paragraph{Translation to English} The approach of leveraging an English-based classifier \texttt{DistillRoBERTa-Emo-EN} as a proxy demonstrated competitive performance as well. Notably, it achieved one of the highest scores for the \texttt{Anger} category, where many other models struggled. Although its precision was lower compared to even the linguistic-based methods, it consistently delivered substantially higher recall. Thus, it can be quite a good baseline for Ukrainian emotional texts detection.

\paragraph{Natural vs Translated Training Data} From Table~\ref{tab:results_synthetic}, we observe that models trained on the original Ukrainian data consistently outperform models tuned on synthetic translated from English training data. However, the latter in some cases achieve higher scores for the \texttt{Anger} class, suggesting---in line with previous observations with the models containing knowledge of English---that English data could be a valuable for augmenting Ukrainian samples for it.

\section{Conclusion}
We introduced \emo---the first manually annotated dataset for emotion detection in Ukrainian texts. The proposed pipeline combines data preprocessing with a two-stage annotation procedure, incorporating multiple quality control measurements to ensure the high quality annotation. We benchmarked a wide range of approaches for the multi-label emotion classification task, demonstrating that although the latest LLMs, such as DeepSeek, achieved the strongest results, more efficient encoder-based models perform competitively. We hope this work also encourages further research on Ukrainian-specific emotion detectors, including ensemble strategies and augmentation with other resource-rich languages resources.

Although the collected Ukrainian dataset is smaller than comparable English resources, its natural, culturally grounded data has already proven more effective than purely cross-lingual transfer approaches. Medium-sized encoder-based Transformers fine-tuned on our dataset achieve performance on par with larger LLMs. We therefore believe that our openly released with all required details data collection pipeline offer a replicable framework for building high-quality and enough in size resources for other underrepresented languages. Finally, we believe that our experimental setup---baselines selection and prompts design---offers a straightforward, extensible evaluation framework for other languages providing a possibility to select a corresponding state-of-the-art approach for text-based emotion analysis.

\section*{Limitations}


While this work introduces \emo as a valuable benchmark for emotion detection in Ukrainian texts, we acknowledge several limitations worth addressing and exploring in future research.

\paragraph{Emotions Labels} The current dataset is restricted to the recognition of basic emotions. More nuanced or implicit emotional states, which often arise in real-world communication, remain outside the scope of this release.

Another challenge is the interpretation of the \texttt{None} label, which can reflect both an absence of emotion or still can be a holder for other emotions rather then listed basic ones. Distinguishing between these two cases is non-trivial and requires deeper investigation.

\paragraph{Emojis as Keywords} The role of non-verbal cues---in particular, the presence of emojis in social media texts---has not been systematically investigated in this work. Emojis can often serve as strong emotion indicators, and future experiments could benefit from incorporating emoji-aware detectors.

\paragraph{Crowdsourcing Platform} Additionally, while the annotation process was performed on a specific crowdsourcing platform---Toloka.ai---we believe that the design of the annotation pipeline is platform-agnostic as annotation guidelines and quality control measures are openly available.

\paragraph{Annotators Subjectivity} Although each instance in the dataset was annotated by five independent annotators, emotions are still highly subjective and culturally sensitive. Increasing annotator overlap, as well as ensuring broader demographic diversity---i.e. Ukrainian speakers from various regions of the country and more diverse age distribution---could further improve label robustness.

\paragraph{Detectors Design} This study focused on evaluating one representative model per classifier type. Future work could explore ensemble methods or hybrid architectures, which have the potential to further enhance performance.

\paragraph{Hyperparameters} Lastly, hyperparameter optimization was explored in a limited setup. More systematic tuning, particularly for prompting strategies (e.g., temperature settings) and fine-tuning, is likely to yield additional improvements.

\section*{Ethics Statement}

We also consider several ethical implications of our work.

During data collection, we made our best to ensure that all contributors were fairly compensated. Clear guidelines and examples were provided to reduce potential ambiguity or emotional strain on the annotators.

All texts in the dataset originate from publicly available sources and were anonymized with totally removed links and any users mentioning to avoid the disclosure of personal or sensitive information. Nonetheless, since the source data comes from social media, there remains a potential for indirect identification through unique expressions or context. We encourage future users of the dataset to handle the material responsibly.

Given the subjective nature of emotions and their cultural grounding, we acknowledge that both annotation and model predictions may reflect current social and cultural biases. This is a general limitation for emotion or other style recognition datasets. We advise the stakeholders of the potential applications to additionally cross-check the models and data for their specific use-cases with corresponding to context adjustments.

Finally, we openly release the annotation guidelines for transparency and reproducibility and encourage future work to continue contribute with various data, including emotions detection, for underrepresented languages.

\section*{Acknowledgments}
We would like to express our gratitude to Toloka.ai platform for their research grant for data annotation. The work was supported by the European Research Council (ERC) through the European Union’s Horizon Europe research and innovation programme (grant agreement No. 101113091) and the German Research Foundation (DFG; grant FR 2829/7-1).

\bibliography{anthology,custom}

\begin{thebibliography}{49}
\providecommand{\natexlab}[1]{#1}

\bibitem[{Abdaoui et~al.(2020)Abdaoui, Pradel, and Sigel}]{abdaoui-etal-2020-load}
Amine Abdaoui, Camille Pradel, and Gr{\'e}goire Sigel. 2020.
\newblock \href {https://doi.org/10.18653/v1/2020.sustainlp-1.16} {Load what you need: Smaller versions of mutililingual {BERT}}.
\newblock In \emph{Proceedings of SustaiNLP: Workshop on Simple and Efficient Natural Language Processing}, pages 119--123, Online. Association for Computational Linguistics.

\bibitem[{AI@Meta(2024)}]{llama3modelcard}
AI@Meta. 2024.
\newblock \href {https://github.com/meta-llama/llama3/blob/main/MODEL_CARD.md} {Llama 3 {M}odel {C}ard}.
\newblock Accessed: 2025-09-18.

\bibitem[{Al-Omari et~al.(2020)Al-Omari, Abdullah, and Shaikh}]{al2020emodet2}
Hani Al-Omari, Malak~A Abdullah, and Samira Shaikh. 2020.
\newblock Emodet2: Emotion detection in english textual dialogue using bert and bilstm models.
\newblock In \emph{2020 11th International Conference on Information and Communication Systems (ICICS)}, pages 226--232. IEEE.

\bibitem[{Barbieri et~al.(2022)Barbieri, Espinosa~Anke, and Camacho-Collados}]{barbieri-etal-2022-xlm}
Francesco Barbieri, Luis Espinosa~Anke, and Jose Camacho-Collados. 2022.
\newblock \href {https://aclanthology.org/2022.lrec-1.27/} {{XLM}-{T}: Multilingual language models in {T}witter for sentiment analysis and beyond}.
\newblock In \emph{Proceedings of the Thirteenth Language Resources and Evaluation Conference}, pages 258--266, Marseille, France. European Language Resources Association.

\bibitem[{Bobrovnyk(2019)}]{bobrovnik2019twt}
Kateryna Bobrovnyk. 2019.
\newblock Automated building and analysis of ukrainian twitter corpus for toxic text detection.
\newblock In \emph{COLINS 2019. Volume II: Workshop}.

\bibitem[{Brauwers and Frasincar(2023)}]{DBLP:journals/csur/BrauwersF23}
Gianni Brauwers and Flavius Frasincar. 2023.
\newblock \href {https://doi.org/10.1145/3503044} {A survey on aspect-based sentiment classification}.
\newblock \emph{{ACM} Comput. Surv.}, 55(4):65:1--65:37.

\bibitem[{Chaplynskyi(2023)}]{chaplynskyi-2023-introducing}
Dmytro Chaplynskyi. 2023.
\newblock \href {https://doi.org/10.18653/v1/2023.unlp-1.1} {Introducing {U}ber{T}ext 2.0: A corpus of {M}odern {U}krainian at scale}.
\newblock In \emph{Proceedings of the Second Ukrainian Natural Language Processing Workshop (UNLP)}, pages 1--10, Dubrovnik, Croatia. Association for Computational Linguistics.

\bibitem[{Chatterjee et~al.(2019)Chatterjee, Narahari, Joshi, and Agrawal}]{chatterjee-etal-2019-semeval}
Ankush Chatterjee, Kedhar~Nath Narahari, Meghana Joshi, and Puneet Agrawal. 2019.
\newblock \href {https://doi.org/10.18653/v1/S19-2005} {{S}em{E}val-2019 task 3: {E}mo{C}ontext contextual emotion detection in text}.
\newblock In \emph{Proceedings of the 13th International Workshop on Semantic Evaluation}, pages 39--48, Minneapolis, Minnesota, USA. Association for Computational Linguistics.

\bibitem[{Conneau et~al.(2019)Conneau, Khandelwal, Goyal, Chaudhary, Wenzek, Guzm{\'{a}}n, Grave, Ott, Zettlemoyer, and Stoyanov}]{DBLP:journals/corr/abs-1911-02116}
Alexis Conneau, Kartikay Khandelwal, Naman Goyal, Vishrav Chaudhary, Guillaume Wenzek, Francisco Guzm{\'{a}}n, Edouard Grave, Myle Ott, Luke Zettlemoyer, and Veselin Stoyanov. 2019.
\newblock \href {https://arxiv.org/abs/1911.02116} {Unsupervised cross-lingual representation learning at scale}.
\newblock \emph{CoRR}, abs/1911.02116.

\bibitem[{Conneau et~al.(2020)Conneau, Khandelwal, Goyal, Chaudhary, Wenzek, Guzm{\'a}n, Grave, Ott, Zettlemoyer, and Stoyanov}]{conneau-etal-2020-unsupervised}
Alexis Conneau, Kartikay Khandelwal, Naman Goyal, Vishrav Chaudhary, Guillaume Wenzek, Francisco Guzm{\'a}n, Edouard Grave, Myle Ott, Luke Zettlemoyer, and Veselin Stoyanov. 2020.
\newblock \href {https://doi.org/10.18653/v1/2020.acl-main.747} {Unsupervised cross-lingual representation learning at scale}.
\newblock In \emph{Proceedings of the 58th Annual Meeting of the Association for Computational Linguistics}, pages 8440--8451, Online. Association for Computational Linguistics.

\bibitem[{Costa{-}juss{\`{a}} et~al.(2022)Costa{-}juss{\`{a}}, Cross, {\c{C}}elebi, Elbayad, Heafield, Heffernan, Kalbassi, Lam, Licht, Maillard, Sun, Wang, Wenzek, Youngblood, Akula, Barrault, Gonzalez, Hansanti, Hoffman, Jarrett, Sadagopan, Rowe, Spruit, Tran, Andrews, Ayan, Bhosale, Edunov, Fan, Gao, Goswami, Guzm{\'{a}}n, Koehn, Mourachko, Ropers, Saleem, Schwenk, and Wang}]{costa2022no}
Marta~R. Costa{-}juss{\`{a}}, James Cross, Onur {\c{C}}elebi, Maha Elbayad, Kenneth Heafield, Kevin Heffernan, Elahe Kalbassi, Janice Lam, Daniel Licht, Jean Maillard, Anna~Y. Sun, Skyler Wang, Guillaume Wenzek, Al~Youngblood, Bapi Akula, Lo{\"{\i}}c Barrault, Gabriel~Mejia Gonzalez, Prangthip Hansanti, John Hoffman, and 19 others. 2022.
\newblock \href {https://doi.org/10.48550/ARXIV.2207.04672} {No language left behind: Scaling human-centered machine translation}.
\newblock \emph{CoRR}, abs/2207.04672.

\bibitem[{Dawid and Skene(1979)}]{dawid1979maximum}
Alexander~Philip Dawid and Allan~M Skene. 1979.
\newblock Maximum likelihood estimation of observer error-rates using the em algorithm.
\newblock \emph{Journal of the Royal Statistical Society: Series C (Applied Statistics)}, 28(1):20--28.

\bibitem[{De~Bruyne(2023)}]{de-bruyne-2023-paradox}
Luna De~Bruyne. 2023.
\newblock \href {https://doi.org/10.18653/v1/2023.wassa-1.40} {The paradox of multilingual emotion detection}.
\newblock In \emph{Proceedings of the 13th Workshop on Computational Approaches to Subjectivity, Sentiment, {\&} Social Media Analysis}, pages 458--466, Toronto, Canada. Association for Computational Linguistics.

\bibitem[{De~Bruyne et~al.(2022)De~Bruyne, Singh, De~Clercq, Lefever, and Hoste}]{de-bruyne-etal-2022-language}
Luna De~Bruyne, Pranaydeep Singh, Orphee De~Clercq, Els Lefever, and Veronique Hoste. 2022.
\newblock \href {https://doi.org/10.18653/v1/2022.mrl-1.7} {How language-dependent is emotion detection? evidence from multilingual {BERT}}.
\newblock In \emph{Proceedings of the 2nd Workshop on Multi-lingual Representation Learning (MRL)}, pages 76--85, Abu Dhabi, United Arab Emirates (Hybrid). Association for Computational Linguistics.

\bibitem[{DeepSeek{-}AI et~al.(2025)DeepSeek{-}AI, Guo, Yang, Zhang, Song, Zhang, Xu, Zhu, Ma, Wang, Bi, Zhang, Yu, Wu, Wu, Gou, Shao, Li, Gao, Liu, Xue, Wang, Wu, Feng, Lu, Zhao, Deng, Zhang, Ruan, Dai, Chen, Ji, Li, Lin, Dai, Luo, Hao, Chen, Li, Zhang, Bao, Xu, Wang, Ding, Xin, Gao, Qu, Li, Guo, Li, Wang, Chen, Yuan, Qiu, Li, Cai, Ni, Liang, Chen, Dong, Hu, Gao, Guan, Huang, Yu, Wang, Zhang, Zhao, Wang, Zhang, Xu, Xia, Zhang, Zhang, Tang, Li, Wang, Li, Tian, Huang, Zhang, Wang, Chen, Du, Ge, Zhang, Pan, Wang, Chen, Jin, Chen, Lu, Zhou, Chen, Ye, Wang, Yu, Zhou, Pan, and Li}]{DBLP:journals/corr/abs-2501-12948}
DeepSeek{-}AI, Daya Guo, Dejian Yang, Haowei Zhang, Junxiao Song, Ruoyu Zhang, Runxin Xu, Qihao Zhu, Shirong Ma, Peiyi Wang, Xiao Bi, Xiaokang Zhang, Xingkai Yu, Yu~Wu, Z.~F. Wu, Zhibin Gou, Zhihong Shao, Zhuoshu Li, Ziyi Gao, and 81 others. 2025.
\newblock \href {https://doi.org/10.48550/ARXIV.2501.12948} {Deepseek-r1: Incentivizing reasoning capability in llms via reinforcement learning}.
\newblock \emph{CoRR}, abs/2501.12948.

\bibitem[{Dementieva et~al.(2023)Dementieva, Babakov, and Panchenko}]{dementieva-etal-2023-detecting}
Daryna Dementieva, Nikolay Babakov, and Alexander Panchenko. 2023.
\newblock \href {https://aclanthology.org/2023.ranlp-1.31/} {Detecting text formality: A study of text classification approaches}.
\newblock In \emph{Proceedings of the 14th International Conference on Recent Advances in Natural Language Processing}, pages 274--284, Varna, Bulgaria. INCOMA Ltd., Shoumen, Bulgaria.

\bibitem[{Dementieva et~al.(2024)Dementieva, Khylenko, Babakov, and Groh}]{dementieva-etal-2024-toxicity}
Daryna Dementieva, Valeriia Khylenko, Nikolay Babakov, and Georg Groh. 2024.
\newblock \href {https://doi.org/10.18653/v1/2024.woah-1.19} {Toxicity classification in {U}krainian}.
\newblock In \emph{Proceedings of the 8th Workshop on Online Abuse and Harms (WOAH 2024)}, pages 244--255, Mexico City, Mexico. Association for Computational Linguistics.

\bibitem[{Dementieva et~al.(2025)Dementieva, Khylenko, and Groh}]{dementieva-etal-2025-cross}
Daryna Dementieva, Valeriia Khylenko, and Georg Groh. 2025.
\newblock \href {https://aclanthology.org/2025.coling-main.97/} {Cross-lingual text classification transfer: The case of {U}krainian}.
\newblock In \emph{Proceedings of the 31st International Conference on Computational Linguistics}, pages 1451--1464, Abu Dhabi, UAE. Association for Computational Linguistics.

\bibitem[{Devlin et~al.(2018)Devlin, Chang, Lee, and Toutanova}]{DBLP:journals/corr/abs-1810-04805}
Jacob Devlin, Ming{-}Wei Chang, Kenton Lee, and Kristina Toutanova. 2018.
\newblock \href {https://arxiv.org/abs/1810.04805} {{BERT:} pre-training of deep bidirectional transformers for language understanding}.
\newblock \emph{CoRR}, abs/1810.04805.

\bibitem[{Ekman et~al.(1999)Ekman, Dalgleish, and Power}]{ekman1999basic}
Paul Ekman, Tim Dalgleish, and M~Power. 1999.
\newblock Basic emotions.
\newblock \emph{San Francisco, USA}.

\bibitem[{Feng et~al.(2022)Feng, Yang, Cer, Arivazhagan, and Wang}]{feng-etal-2022-language}
Fangxiaoyu Feng, Yinfei Yang, Daniel Cer, Naveen Arivazhagan, and Wei Wang. 2022.
\newblock \href {https://doi.org/10.18653/v1/2022.acl-long.62} {Language-agnostic {BERT} sentence embedding}.
\newblock In \emph{Proceedings of the 60th Annual Meeting of the Association for Computational Linguistics (Volume 1: Long Papers)}, pages 878--891, Dublin, Ireland. Association for Computational Linguistics.

\bibitem[{Imani et~al.(2023)Imani, Lin, Kargaran, Severini, Jalili~Sabet, Kassner, Ma, Schmid, Martins, Yvon, and Sch{\"u}tze}]{imanigooghari-etal-2023-glot500}
Ayyoob Imani, Peiqin Lin, Amir~Hossein Kargaran, Silvia Severini, Masoud Jalili~Sabet, Nora Kassner, Chunlan Ma, Helmut Schmid, Andr{\'e} Martins, Fran{\c{c}}ois Yvon, and Hinrich Sch{\"u}tze. 2023.
\newblock \href {https://doi.org/10.18653/v1/2023.acl-long.61} {Glot500: Scaling multilingual corpora and language models to 500 languages}.
\newblock In \emph{Proceedings of the 61st Annual Meeting of the Association for Computational Linguistics (Volume 1: Long Papers)}, pages 1082--1117, Toronto, Canada. Association for Computational Linguistics.

\bibitem[{Jiang et~al.(2023)Jiang, Sablayrolles, Mensch, Bamford, Chaplot, de~Las~Casas, Bressand, Lengyel, Lample, Saulnier, Lavaud, Lachaux, Stock, Scao, Lavril, Wang, Lacroix, and Sayed}]{Jiang2023Mistral7}
Albert~Q. Jiang, Alexandre Sablayrolles, Arthur Mensch, Chris Bamford, Devendra~Singh Chaplot, Diego de~Las~Casas, Florian Bressand, Gianna Lengyel, Guillaume Lample, Lucile Saulnier, L{\'{e}}lio~Renard Lavaud, Marie{-}Anne Lachaux, Pierre Stock, Teven~Le Scao, Thibaut Lavril, Thomas Wang, Timoth{\'{e}}e Lacroix, and William~El Sayed. 2023.
\newblock \href {https://doi.org/10.48550/ARXIV.2310.06825} {Mistral 7b}.
\newblock \emph{CoRR}, abs/2310.06825.

\bibitem[{Kravchenko et~al.(2025)Kravchenko, Paniv, and Drushchak}]{kravchenko-etal-2025-ualign}
Andrian Kravchenko, Yurii Paniv, and Nazarii Drushchak. 2025.
\newblock \href {https://doi.org/10.18653/v1/2025.unlp-1.4} {{UA}lign: {LLM} alignment benchmark for the {U}krainian language}.
\newblock In \emph{Proceedings of the Fourth Ukrainian Natural Language Processing Workshop (UNLP 2025)}, pages 36--44, Vienna, Austria (online). Association for Computational Linguistics.

\bibitem[{Kumar et~al.(2023)Kumar, Pathania, and Raman}]{DBLP:journals/apin/KumarPR23}
Puneet Kumar, Kshitij Pathania, and Balasubramanian Raman. 2023.
\newblock \href {https://doi.org/10.1007/S10489-022-04046-6} {Zero-shot learning based cross-lingual sentiment analysis for sanskrit text with insufficient labeled data}.
\newblock \emph{Appl. Intell.}, 53(9):10096--10113.

\bibitem[{Kumar et~al.(2022)Kumar, Shrimal, Akhtar, and Chakraborty}]{DBLP:journals/kbs/KumarSA022}
Shivani Kumar, Anubhav Shrimal, Md.~Shad Akhtar, and Tanmoy Chakraborty. 2022.
\newblock \href {https://doi.org/10.1016/J.KNOSYS.2021.108112} {Discovering emotion and reasoning its flip in multi-party conversations using masked memory network and transformer}.
\newblock \emph{Knowl. Based Syst.}, 240:108112.

\bibitem[{Kusal et~al.(2023)Kusal, Patil, Choudrie, Kotecha, Vora, and Pappas}]{DBLP:journals/air/KusalPCKVP23}
Sheetal Kusal, Shruti Patil, Jyoti Choudrie, Ketan Kotecha, Deepali~Rahul Vora, and Ilias~O. Pappas. 2023.
\newblock \href {https://doi.org/10.1007/S10462-023-10509-0} {A systematic review of applications of natural language processing and future challenges with special emphasis in text-based emotion detection}.
\newblock \emph{Artif. Intell. Rev.}, 56(12):15129--15215.

\bibitem[{Martins et~al.(2024)Martins, Fernandes, Alves, Guerreiro, Rei, Alves, Pombal, Farajian, Faysse, Klimaszewski, Colombo, Haddow, de~Souza, Birch, and Martins}]{DBLP:journals/corr/abs-2409-16235}
Pedro~Henrique Martins, Patrick Fernandes, Jo{\~{a}}o Alves, Nuno~Miguel Guerreiro, Ricardo Rei, Duarte~M. Alves, Jos{\'{e}} Pombal, M.~Amin Farajian, Manuel Faysse, Mateusz Klimaszewski, Pierre Colombo, Barry Haddow, Jos{\'{e}} G.~C. de~Souza, Alexandra Birch, and Andr{\'{e}} F.~T. Martins. 2024.
\newblock \href {https://doi.org/10.48550/ARXIV.2409.16235} {Eurollm: Multilingual language models for europe}.
\newblock \emph{CoRR}, abs/2409.16235.

\bibitem[{Mohammad(2023)}]{mohammad-2023-best}
Saif Mohammad. 2023.
\newblock \href {https://doi.org/10.18653/v1/2023.findings-eacl.136} {Best practices in the creation and use of emotion lexicons}.
\newblock In \emph{Findings of the Association for Computational Linguistics: EACL 2023}, pages 1825--1836, Dubrovnik, Croatia. Association for Computational Linguistics.

\bibitem[{Mohammad et~al.(2018)Mohammad, Bravo-Marquez, Salameh, and Kiritchenko}]{mohammad-etal-2018-semeval}
Saif Mohammad, Felipe Bravo-Marquez, Mohammad Salameh, and Svetlana Kiritchenko. 2018.
\newblock \href {https://doi.org/10.18653/v1/S18-1001} {{S}em{E}val-2018 task 1: Affect in tweets}.
\newblock In \emph{Proceedings of the 12th International Workshop on Semantic Evaluation}, pages 1--17, New Orleans, Louisiana. Association for Computational Linguistics.

\bibitem[{Mohammad(2022)}]{mohammad-2022-ethics-sheet}
Saif~M. Mohammad. 2022.
\newblock \href {https://doi.org/10.1162/coli_a_00433} {Ethics sheet for automatic emotion recognition and sentiment analysis}.
\newblock \emph{Computational Linguistics}, 48(2):239--278.

\bibitem[{Muhammad et~al.(2025{\natexlab{a}})Muhammad, Ousidhoum, Abdulmumin, Wahle, Ruas, Beloucif, de~Kock, Surange, Teodorescu, Ahmad, Adelani, Aji, Ali, Alimova, Araujo, Babakov, Baes, Bucur, Bukula, Cao, Tufi{\~n}o, Chevi, Chukwuneke, Ciobotaru, Dementieva, Gadanya, Geislinger, Gipp, Hourrane, Ignat, Lawan, Mabuya, Mahendra, Marivate, Panchenko, Piper, Ferreira, Protasov, Rutunda, Shrivastava, Udrea, Wanzare, Wu, Wunderlich, Zhafran, Zhang, Zhou, and Mohammad}]{muhammad-etal-2025-brighter}
Shamsuddeen~Hassan Muhammad, Nedjma Ousidhoum, Idris Abdulmumin, Jan~Philip Wahle, Terry Ruas, Meriem Beloucif, Christine de~Kock, Nirmal Surange, Daniela Teodorescu, Ibrahim~Said Ahmad, David~Ifeoluwa Adelani, Alham~Fikri Aji, Felermino D. M.~A. Ali, Ilseyar Alimova, Vladimir Araujo, Nikolay Babakov, Naomi Baes, Ana-Maria Bucur, Andiswa Bukula, and 29 others. 2025{\natexlab{a}}.
\newblock \href {https://doi.org/10.18653/v1/2025.acl-long.436} {{BRIGHTER}: {BRI}dging the gap in human-annotated textual emotion recognition datasets for 28 languages}.
\newblock In \emph{Proceedings of the 63rd Annual Meeting of the Association for Computational Linguistics (Volume 1: Long Papers)}, pages 8895--8916, Vienna, Austria. Association for Computational Linguistics.

\bibitem[{Muhammad et~al.(2025{\natexlab{b}})Muhammad, Ousidhoum, Abdulmumin, Yimam, Wahle, Lima~Ruas, Beloucif, De~Kock, Belay, Ahmad, Surange, Teodorescu, Adelani, Aji, Ali, Araujo, Ayele, Ignat, Panchenko, Zhou, and Mohammad}]{muhammad-etal-2025-semeval}
Shamsuddeen~Hassan Muhammad, Nedjma Ousidhoum, Idris Abdulmumin, Seid~Muhie Yimam, Jan~Philip Wahle, Terry Lima~Ruas, Meriem Beloucif, Christine De~Kock, Tadesse~Destaw Belay, Ibrahim~Said Ahmad, Nirmal Surange, Daniela Teodorescu, David~Ifeoluwa Adelani, Alham~Fikri Aji, Felermino Dario~Mario Ali, Vladimir Araujo, Abinew~Ali Ayele, Oana Ignat, Alexander Panchenko, and 2 others. 2025{\natexlab{b}}.
\newblock \href {https://aclanthology.org/2025.semeval-1.327/} {{S}em{E}val-2025 task 11: Bridging the gap in text-based emotion detection}.
\newblock In \emph{Proceedings of the 19th International Workshop on Semantic Evaluation (SemEval-2025)}, pages 2558--2569, Vienna, Austria. Association for Computational Linguistics.

\bibitem[{{\"O}hman et~al.(2020){\"O}hman, P{\`a}mies, Kajava, and Tiedemann}]{ohman-etal-2020-xed}
Emily {\"O}hman, Marc P{\`a}mies, Kaisla Kajava, and J{\"o}rg Tiedemann. 2020.
\newblock \href {https://doi.org/10.18653/v1/2020.coling-main.575} {{XED}: A multilingual dataset for sentiment analysis and emotion detection}.
\newblock In \emph{Proceedings of the 28th International Conference on Computational Linguistics}, pages 6542--6552, Barcelona, Spain (Online). International Committee on Computational Linguistics.

\bibitem[{Oliinyk and Matviichuk(2023)}]{oliinyk2023low}
V~Oliinyk and I~Matviichuk. 2023.
\newblock \href {https://ela.kpi.ua/items/301aad2e-df26-49e8-aac6-13eeb51107dd} {Low-resource text classification using cross-lingual models for bullying detection in the ukrainian language}.
\newblock \emph{Adaptive systems of automatic control: interdepartmental scientific and technical collection, 2023, 1 (42)}.

\bibitem[{Pfeiffer et~al.(2020)Pfeiffer, R{\"u}ckl{\'e}, Poth, Kamath, Vuli{\'c}, Ruder, Cho, and Gurevych}]{pfeiffer-etal-2020-adapterhub}
Jonas Pfeiffer, Andreas R{\"u}ckl{\'e}, Clifton Poth, Aishwarya Kamath, Ivan Vuli{\'c}, Sebastian Ruder, Kyunghyun Cho, and Iryna Gurevych. 2020.
\newblock \href {https://doi.org/10.18653/v1/2020.emnlp-demos.7} {{A}dapter{H}ub: A framework for adapting transformers}.
\newblock In \emph{Proceedings of the 2020 Conference on Empirical Methods in Natural Language Processing: System Demonstrations}, pages 46--54, Online. Association for Computational Linguistics.

\bibitem[{Plaza~del Arco et~al.(2020)Plaza~del Arco, Strapparava, Urena~Lopez, and Martin}]{plaza-del-arco-etal-2020-emoevent}
Flor~Miriam Plaza~del Arco, Carlo Strapparava, L.~Alfonso Urena~Lopez, and Maite Martin. 2020.
\newblock \href {https://aclanthology.org/2020.lrec-1.186/} {{E}mo{E}vent: A multilingual emotion corpus based on different events}.
\newblock In \emph{Proceedings of the Twelfth Language Resources and Evaluation Conference}, pages 1492--1498, Marseille, France. European Language Resources Association.

\bibitem[{Rivi{\`{e}}re et~al.(2024)Rivi{\`{e}}re, Pathak, Sessa, Hardin, Bhupatiraju, Hussenot, Mesnard, Shahriari, Ram{\'{e}}, Ferret, Liu, Tafti, Friesen, Casbon, Ramos, Kumar, Lan, Jerome, Tsitsulin, Vieillard, Stanczyk, Girgin, Momchev, Hoffman, Thakoor, Grill, Neyshabur, Bachem, Walton, Severyn, Parrish, Ahmad, Hutchison, Abdagic, Carl, Shen, Brock, Coenen, Laforge, Paterson, Bastian, Piot, Wu, Royal, Chen, Kumar, Perry, Welty, Choquette{-}Choo, Sinopalnikov, Weinberger, Vijaykumar, Rogozinska, Herbison, Bandy, Wang, Noland, Moreira, Senter, Eltyshev, Visin, Rasskin, Wei, Cameron, Martins, Hashemi, Klimczak{-}Plucinska, Batra, Dhand, Nardini, Mein, Zhou, Svensson, Stanway, Chan, Zhou, Carrasqueira, Iljazi, Becker, Fernandez, van Amersfoort, Gordon, Lipschultz, Newlan, Ji, Mohamed, Badola, Black, Millican, McDonell, Nguyen, Sodhia, Greene, Sj{\"{o}}sund, Usui, Sifre, Heuermann, Lago, and McNealus}]{DBLP:journals/corr/abs-2408-00118}
Morgane Rivi{\`{e}}re, Shreya Pathak, Pier~Giuseppe Sessa, Cassidy Hardin, Surya Bhupatiraju, L{\'{e}}onard Hussenot, Thomas Mesnard, Bobak Shahriari, Alexandre Ram{\'{e}}, Johan Ferret, Peter Liu, Pouya Tafti, Abe Friesen, Michelle Casbon, Sabela Ramos, Ravin Kumar, Charline~Le Lan, Sammy Jerome, Anton Tsitsulin, and 80 others. 2024.
\newblock \href {https://doi.org/10.48550/ARXIV.2408.00118} {Gemma 2: Improving open language models at a practical size}.
\newblock \emph{CoRR}, abs/2408.00118.

\bibitem[{Saini et~al.(2024)Saini, Chernodub, Raheja, and Kulkarni}]{saini-etal-2024-spivavtor}
Aman Saini, Artem Chernodub, Vipul Raheja, and Vivek Kulkarni. 2024.
\newblock \href {https://aclanthology.org/2024.unlp-1.12/} {Spivavtor: An instruction tuned {U}krainian text editing model}.
\newblock In \emph{Proceedings of the Third Ukrainian Natural Language Processing Workshop (UNLP) @ LREC-COLING 2024}, pages 95--108, Torino, Italia. ELRA and ICCL.

\bibitem[{Shynkarov et~al.(2025)Shynkarov, Solopova, and Schmitt}]{shynkarov-etal-2025-improving}
Yurii Shynkarov, Veronika Solopova, and Vera Schmitt. 2025.
\newblock \href {https://doi.org/10.18653/v1/2025.unlp-1.18} {Improving sentiment analysis for {U}krainian social media code-switching data}.
\newblock In \emph{Proceedings of the Fourth Ukrainian Natural Language Processing Workshop (UNLP 2025)}, pages 179--193, Vienna, Austria (online). Association for Computational Linguistics.

\bibitem[{Team(2025)}]{qwen3technicalreport}
Qwen Team. 2025.
\newblock \href {https://arxiv.org/abs/2505.09388} {Qwen3 technical report}.
\newblock \emph{Preprint}, arXiv:2505.09388.

\bibitem[{Tiedemann(2012)}]{tiedemann-2012-parallel}
J{\"o}rg Tiedemann. 2012.
\newblock \href {https://aclanthology.org/L12-1246/} {Parallel data, tools and interfaces in {OPUS}}.
\newblock In \emph{Proceedings of the Eighth International Conference on Language Resources and Evaluation ({LREC}`12)}, pages 2214--2218, Istanbul, Turkey. European Language Resources Association (ELRA).

\bibitem[{{\"{U}}st{\"{u}}n et~al.(2024){\"{U}}st{\"{u}}n, Aryabumi, Yong, Ko, D'souza, Onilude, Bhandari, Singh, Ooi, Kayid, Vargus, Blunsom, Longpre, Muennighoff, Fadaee, Kreutzer, and Hooker}]{DBLP:conf/acl/UstunAYKDOBSOKV24}
Ahmet {\"{U}}st{\"{u}}n, Viraat Aryabumi, Zheng~Xin Yong, Wei{-}Yin Ko, Daniel D'souza, Gbemileke Onilude, Neel Bhandari, Shivalika Singh, Hui{-}Lee Ooi, Amr Kayid, Freddie Vargus, Phil Blunsom, Shayne Longpre, Niklas Muennighoff, Marzieh Fadaee, Julia Kreutzer, and Sara Hooker. 2024.
\newblock \href {https://doi.org/10.18653/V1/2024.ACL-LONG.845} {Aya model: An instruction finetuned open-access multilingual language model}.
\newblock In \emph{Proceedings of the 62nd Annual Meeting of the Association for Computational Linguistics (Volume 1: Long Papers), {ACL} 2024, Bangkok, Thailand, August 11-16, 2024}, pages 15894--15939. Association for Computational Linguistics.

\bibitem[{Vaswani et~al.(2017)Vaswani, Shazeer, Parmar, Uszkoreit, Jones, Gomez, Kaiser, and Polosukhin}]{DBLP:conf/nips/VaswaniSPUJGKP17}
Ashish Vaswani, Noam Shazeer, Niki Parmar, Jakob Uszkoreit, Llion Jones, Aidan~N. Gomez, Lukasz Kaiser, and Illia Polosukhin. 2017.
\newblock \href {https://proceedings.neurips.cc/paper/2017/hash/3f5ee243547dee91fbd053c1c4a845aa-Abstract.html} {Attention is all you need}.
\newblock In \emph{Advances in Neural Information Processing Systems 30: Annual Conference on Neural Information Processing Systems 2017, December 4-9, 2017, Long Beach, CA, {USA}}, pages 5998--6008.

\bibitem[{Wang et~al.(2024{\natexlab{a}})Wang, Yang, Huang, Yang, Majumder, and Wei}]{DBLP:journals/corr/abs-2402-05672}
Liang Wang, Nan Yang, Xiaolong Huang, Linjun Yang, Rangan Majumder, and Furu Wei. 2024{\natexlab{a}}.
\newblock \href {https://doi.org/10.48550/ARXIV.2402.05672} {Multilingual {E5} text embeddings: {A} technical report}.
\newblock \emph{CoRR}, abs/2402.05672.

\bibitem[{Wang et~al.(2024{\natexlab{b}})Wang, Wang, Han, Wang, Chen, Zhang, Pan, and Nguyen}]{wang-etal-2024-knowledge}
Yuqi Wang, Zimu Wang, Nijia Han, Wei Wang, Qi~Chen, Haiyang Zhang, Yushan Pan, and Anh Nguyen. 2024{\natexlab{b}}.
\newblock \href {https://doi.org/10.18653/v1/2024.wassa-1.45} {Knowledge distillation from monolingual to multilingual models for intelligent and interpretable multilingual emotion detection}.
\newblock In \emph{Proceedings of the 14th Workshop on Computational Approaches to Subjectivity, Sentiment, {\&} Social Media Analysis}, pages 470--475, Bangkok, Thailand. Association for Computational Linguistics.

\bibitem[{Wei et~al.(2022)Wei, Tay, Bommasani, Raffel, Zoph, Borgeaud, Yogatama, Bosma, Zhou, Metzler, Chi, Hashimoto, Vinyals, Liang, Dean, and Fedus}]{DBLP:journals/tmlr/WeiTBRZBYBZMCHVLDF22}
Jason Wei, Yi~Tay, Rishi Bommasani, Colin Raffel, Barret Zoph, Sebastian Borgeaud, Dani Yogatama, Maarten Bosma, Denny Zhou, Donald Metzler, Ed~H. Chi, Tatsunori Hashimoto, Oriol Vinyals, Percy Liang, Jeff Dean, and William Fedus. 2022.
\newblock \href {https://openreview.net/forum?id=yzkSU5zdwD} {Emergent abilities of large language models}.
\newblock \emph{Trans. Mach. Learn. Res.}, 2022.

\bibitem[{Yukhymenko et~al.(2025)Yukhymenko, Alexandrov, and Vechev}]{MamayLM}
Hanna Yukhymenko, Anton Alexandrov, and Martin Vechev. 2025.
\newblock \href {https://huggingface.co/blog/INSAIT-Institute/mamaylm} {Mamaylm: An efficient state-of-the-art ukrainian llm}.

\bibitem[{Zalutska et~al.(2023)Zalutska, Molchanova, Sobko, Mazurets, Pasichnyk, Barmak, and Krak}]{DBLP:conf/colins/ZalutskaMSMPBK23}
Olha Zalutska, Maryna Molchanova, Olena Sobko, Olexander Mazurets, Oleksandr Pasichnyk, Olexander Barmak, and Iurii Krak. 2023.
\newblock \href {https://ceur-ws.org/Vol-3387/paper26.pdf} {Method for sentiment analysis of ukrainian-language reviews in e-commerce using roberta neural network}.
\newblock In \emph{Proceedings of the 7th International Conference on Computational Linguistics and Intelligent Systems. Volume {I:} Machine Learning Workshop, Kharkiv, Ukraine, April 20-21, 2023}, volume 3387 of \emph{{CEUR} Workshop Proceedings}, pages 344--356. CEUR-WS.org.

\end{thebibliography}

\onecolumn
\appendix

\section{\emo Released Datasets and Models}
\label{sec:our_resources}
We release all the collected data and fine-tuned best-performing classifier for public usage for further research purposes and usage for social good. We opensource the complete annotation results---both binary and intensity labels---along with the full record of annotator responses for each text:

\begin{table}[ht!]
\centering
\footnotesize
\begin{tabular}{p{3cm}p{5cm}p{7cm}}
\toprule
Resource & License & Homepage  \\ 
\midrule
Dataset Binary Labels & CC BY 4.0 & \href{https://huggingface.co/datasets/ukr-detect/ukr-emotions-binary}{https://huggingface.co/datasets/ukr-detect/ukr-emotions-binary} \\
Dataset Intensity Labels & CC BY 4.0 & \href{https://huggingface.co/datasets/ukr-detect/ukr-emotions-intensity}{https://huggingface.co/datasets/ukr-detect/ukr-emotions-intensity} \\
Dataset Per Annotator Labels & CC BY 4.0 & \href{https://huggingface.co/datasets/ukr-detect/ukr-emotions-per-annotator}{https://huggingface.co/datasets/ukr-detect/ukr-emotions-per-annotator} \\
SOTA Ukrainian Emotions Classifier & OpenRail++ & \href{https://huggingface.co/ukr-detect/ukr-emotions-classifier}{https://huggingface.co/ukr-detect/ukr-emotions-classifier} \\
\bottomrule
\end{tabular}
\caption{Overview of the licenses associated with our published resources.}
\label{tab:resources-license}
\end{table}

The full project page together with the annotation instructions details, interfaces, and experiments code can be found at the Github page: \href{https://github.com/dardem/emobench-ua}{https://github.com/dardem/emobench-ua}.

\section{Licensing of Resources}
\label{sec:app_licenses}

Below is an overview of the licenses associated with each resource used in this work (Table~\ref{tab:overview-license}).

\begin{table}[ht!]
\centering
\footnotesize
\begin{tabular}{p{3cm}p{5cm}p{7cm}}
\toprule
Resource & License & Homepage  \\ 
\midrule
Ukrainian Tweets Dataset & CC BY 4.0 & \href{https://ena.lpnu.ua:8443/server/api/core/bitstreams/c4c645c1-f465-4895-98dd-765f862cf186/content}{https://ena.lpnu.ua:8443/server/api/core/\newline bitstreams/c4c645c1-f465-4895-98dd-765f862cf186/content} \\
Ukrainian Toxicity Classifier & OpenRail++ & \href{https://huggingface.co/ukr-detect}{https://huggingface.co/ukr-detect} \\
Emotion Lexicon & The lexicon is made freely available for research, and has been commercially licensed to companies for a small fee & \href{https://saifmohammad.com/WebPages/NRC-Emotion-Lexicon.htm}{https://saifmohammad.com/WebPages/NRC-Emotion-Lexicon.htm}\\
mBERT & Apache-2.0 & \href{https://huggingface.co/google-bert}{https://huggingface.co/google-bert}\\
Geotrend-BERT & Apache-2.0 & \href{https://huggingface.co/Geotrend/bert-base-uk-cased}{https://huggingface.co/Geotrend/bert-base-uk-cased}\\
XLM-RoBERTa & MIT & \href{https://huggingface.co/FacebookAI}{https://huggingface.co/FacebookAI} \\
UKR-RoBERTa & MIT & \href{https://github.com/youscan/language-models}{https://github.com/youscan/language-models}\\
Twitter-XLM-RoBERTa & Apache-2.0 & \href{https://aclanthology.org/2022.lrec-1.27}{https://aclanthology.org/2022.lrec-1.27} \\
Glot500 & CC BY 4.0 & \href{https://aclanthology.org/2023.acl-long.61}{https://aclanthology.org/2023.acl-long.61}\\
LaBSE & Apache-2.0 & \href{https://huggingface.co/sentence-transformers/LaBSE}{https://huggingface.co/sentence-transformers/LaBSE}\\
e5 & MIT & \href{https://huggingface.co/intfloat}{https://huggingface.co/intfloat} \\
NLLB & CC BY NC 4.0 & \href{https://huggingface.co/facebook/nllb-200-distilled-600M}{https://huggingface.co/facebook/nllb-200-distilled-600M} \\
EuroLLM & Apache-2.0 & \href{https://huggingface.co/utter-project/EuroLLM-1.7B-Instruct}{https://huggingface.co/utter-project/EuroLLM-1.7B-Instruct}\\
Spivavtor & CC BY 4.0 & \href{https://huggingface.co/collections/grammarly/spivavtor-660744ab14fdf5e925592dc7}{https://huggingface.co/collections/grammarly/\newline spivavtor-660744ab14fdf5e925592dc7}\\
MamayLM & Gemma License & \href{https://huggingface.co/collections/INSAIT-Institute/mamaylm-gemma-2-68080b895a949a52b474d5de}{https://huggingface.co/collections/INSAIT-Institute/mamaylm-gemma-2-68080b895a949a52b474d5de} \\
Mistral7B & Apache-2.0 & \href{https://huggingface.co/mistralai}{https://huggingface.co/mistralai} \\
Mixstral8x7B & Apache-2.0 & \href{https://huggingface.co/mistralai}{https://huggingface.co/mistralai} \\
LLaMa3 & llama3 & \href{https://huggingface.co/meta-llama}{https://huggingface.co/meta-llama} \\
Qwen3 & Apache-2.0 & \href{https://huggingface.co/collections/Qwen/qwen3-67dd247413f0e2e4f653967f}{https://huggingface.co/collections/Qwen/qwen3-67dd247413f0e2e4f653967f} \\
DeepSeek & MIT & \href{https://huggingface.co/collections/deepseek-ai/deepseek-r1-678e1e131c0169c0bc89728d}{https://huggingface.co/collections/deepseek-ai/deepseek-r1-678e1e131c0169c0bc89728d}\\
\bottomrule
\end{tabular}
\caption{Overview of the licenses associated with each resource utilized in this work for experiments.}
\label{tab:overview-license}
\end{table}

The licenses associated with the models and datasets utilized in this study are consistent with the intended use of conducting academic research on various NLP application for positive impact.

\section{Usage of AI Assistants}

During this study, AI assistant was utilized in the writing process. ChatGPT was employed for paraphrasing and improving clarity throughout the paper’s formulation. We also utilized DeepL\footnote{\href{https://www.deepl.com}{https://www.deepl.com}} to translate the examples in Ukrainian into English followed by the human manual check and adjustments.


\section{Instructions \& Interface}
\label{sec:app_instructions}

\subsection{Ukrainian (original)}

In this section, we provide the Instructions for both annotation projects as well as interface in Ukrainian. 

\begin{tcolorbox}[breakable, colback=blue!10!white, colframe=blue!50!black, title=\texttt{Main Instructions for the First Project: Fear, Surprise, Disgust}]
\selectlanguage{ukrainian}

Виберіть одну або кілька емоцій та їх інтенсивність у тексті. Якщо в тексті немає ніяких емоцій або є емоції не представлені в списку виберіть варіант - ”Немає емоцій / інші емоції”.
\\\\
\textbf{Приклади}
\\\\
\textbf{Страх}
\\
Низька проява \\
\hspace*{0.5cm} Що, як це ніколи не закінчиться?
\\\\
Нормальна проява \\
\hspace*{0.5cm} Мені дуже страшно залишатися тут одному… 
\\\\
Iнтесивна проява \\
\hspace*{0.5cm} Боже, який це жах і як же це страшно!!!
\\\\
\textbf{Здивування}
\\
Низька проява\\
\hspace*{0.5cm} Це було несподівано
\\\\
Нормальна проява\\
\hspace*{0.5cm} Це вражає! Я в захваті!
\\\\
Iнтесивна проява \\
Ваууу, який неймовірний поворот подій!!!
\\\\
\textbf{Огида}
\\
Низька проява\\
\hspace*{0.5cm} Щось мене трохи нудить від цього запаху.
\\\\
Нормальна проява \\
\hspace*{0.5cm} Фу, це просто огидно!
\\\\
Iнтесивна проява \\
\hspace*{0.5cm} Мені стає погано від однієї лише думки про це 
\\\\
\textbf{Приклади з декількома емоціями} \\
\hspace*{0.5cm} Ти ще куриш на ходу в таку погоду. – здивування, огида \\
\hspace*{0.5cm} Я боюсь, що це все виявиться п'яними розмовами. – огида, страх \\
\hspace*{0.5cm} Я не можу повірити, що це дійсно сталося! Це так страшно! – здивування, страх \\
\hspace*{0.5cm} Як це можливо? Я боюся навіть уявити, що буде далі! – здивування, страх \\
\hspace*{0.5cm} Я не можу повірити, що хтось може їсти таке! Це жахливо! – огида, здивування \\
\\\\
\textbf{Немає емоцій / інші емоції}
\\\\
\textbf{Немає емоцій}
\\
\hspace*{0.5cm} Сьогодні вранці йшов дощ. \\
\hspace*{0.5cm} Він прочитав книгу за два дні. \\
\hspace*{0.5cm} Я бачив її вчора на вулиці.
\\\\
\textbf{Iнші емоції}
 \\
\hspace*{0.5cm} Я вкрай роздратований цим безладом! \\
\hspace*{0.5cm} Моє серце розривається від болю :(  \\
\hspace*{0.5cm} Нарешті ми це зробили :):) я просто на сьомому небі від щастя! 

\end{tcolorbox}

\begin{tcolorbox}[breakable, colback=blue!10!white, colframe=blue!50!black, title=\texttt{Main Instructions for the Second Project: Joy, Sadness, Anger}]
\selectlanguage{ukrainian}

Виберіть одну або кілька емоцій та їх інтенсивність у тексті. Якщо в тексті немає ніяких емоцій або є емоції не представлені в списку виберіть варіант - ”Немає емоцій / інші емоції”.
\\\\
Приклад
\\\\
Радість
\\
Низька проява \\
\hspace*{0.5cm} Твоя усмішка робить мій день.
\\\\
Нормальна проява \\
\hspace*{0.5cm} Це один з найкращих подарунків, який я коли-небудь отримував. \\ 
\hspace*{0.5cm} Це було дуже весело та чудово, наш відпочинок вдався!! \\
\\\\
Iнтесивна проява \\
\hspace*{0.5cm} Нарешті ми це зробили!!!!! я просто на сьомому небі від щастя! \\
\hspace*{0.5cm} Ми виграли!!! :):) Я не можу повірити, що це сталося! \\
\\\\
Сум
\\
Низька проява \\
\hspace*{0.5cm} Цей день був важкий для мене. \\
\\\\
Нормальна проява\\
\hspace*{0.5cm} Я не можу повірити, що це сталося з нами…
\\\\
Iнтесивна проява\\
\hspace*{0.5cm} Моє серце розривається від болю :(( 
\\\\
Гнів
\\
Низька проява \\
\hspace*{0.5cm} Це мене бісить 
\\\\
Нормальна проява \\
\hspace*{0.5cm} Я вкрай роздратований цим безладом! 
\\\\
Iнтесивна проява \\
\hspace*{0.5cm} Це абсолютно неприпустимо!!! 
\\\\
Приклади з декількома емоціями \\
\hspace*{0.5cm} Нарешті ми знайшли ідеальне місце для відпочинку, і це навіть краще, ніж я міг собі уявити! – радість, здивування \\
\hspace*{0.5cm} Вау, як неочікувано, це найкращий подарунок, який я коли-небудь отримував! –  радість, здивування \\
\hspace*{0.5cm} Мені приємно, що ти прийшов, але ти капець як запізнився!!! – радість, гнів \\
\hspace*{0.5cm} Мені важко прийняти, що все закінчилося саме так, і я злюся на тебе за це. – сум, гнів \\
\hspace*{0.5cm} Це так прикро і гнітюче, що наші відносини закінчилися через твою брехню! – гнів, сум 
\\\\
Немає емоцій / інші емоції
\\\\
Немає емоцій \\
\hspace*{0.5cm} Сьогодні вранці йшов дощ. \\
\hspace*{0.5cm} Він прочитав книгу за два дні. \\
\hspace*{0.5cm} Я бачив її вчора на вулиці.
\end{tcolorbox}

\begin{figure}[h!]
    \centering
    \includegraphics[width=0.5\linewidth]{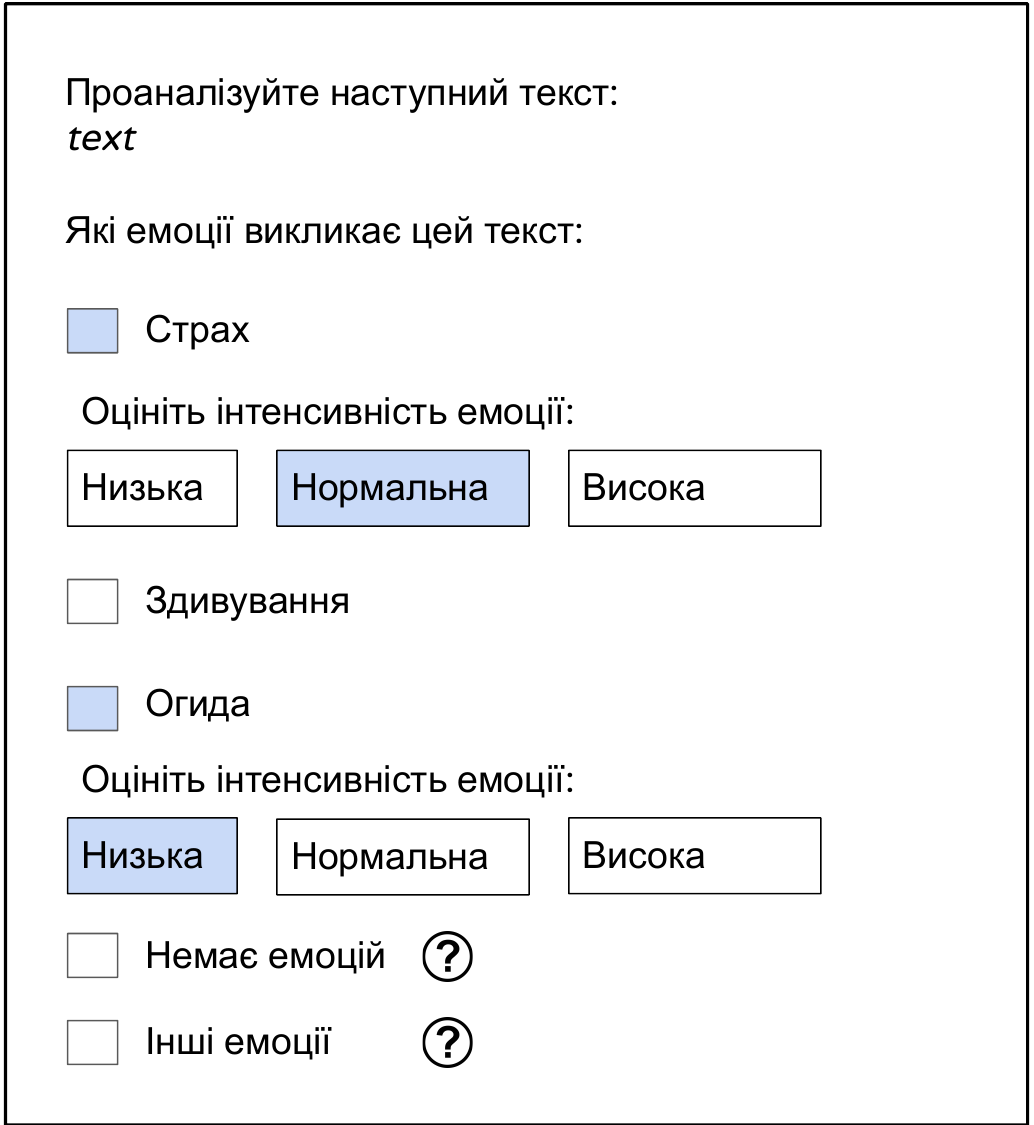}
    \caption{Annotation Interface illustration in original Ukrainian.}
    \label{fig:interface_uk}
\end{figure}

\subsection{English (translated)}

\begin{tcolorbox}[breakable, colback=blue!10!white, colframe=blue!50!black, title=\texttt{Main Instructions for the First Project: Fear, Surprise, Disgust}]
\selectlanguage{english}

Select one or more emotions and their intensity in the text. If there are no emotions in the text or if there are emotions not represented in the list, select the No emotions / other emotions option.
\\\\
\textbf{Examples}
\\\\
\textbf{Fear}
\\
Low\\
\hspace*{0.5cm} What if it never ends?
\\\\
Normal \\
\hspace*{0.5cm} I am very scared to stay here alone...
\\\\
High \\
\hspace*{0.5cm} My God, what a horror and how scary it is!!!
\\\\
\textbf{Surprise}
\\
Low\\
\hspace*{0.5cm} It was unexpected
\\\\
Normal\\
\hspace*{0.5cm} It's amazing! I'm thrilled!
\\\\
High \\
Wow, what an incredible turn of events!!!
\\\\
\textbf{Disgust}
\\
Low\\
\hspace*{0.5cm} 
This smell makes me a little nauseous.
\\\\
Normal \\
\hspace*{0.5cm} Ew, that's just disgusting!
\\\\
High\\
\hspace*{0.5cm} 
I feel sick just thinking about it 
\\\\
\textbf{Examples with multiple emotions} \\
\hspace*{0.5cm} You're still smoking on the go in this weather. - surprise, disgust\\
\hspace*{0.5cm} I'm afraid it will all turn out to be drunken talk. - disgust, fear \\
\hspace*{0.5cm} I can't believe this really happened! It's so scary! - surprise, fear \\
\hspace*{0.5cm} How is this possible? I'm afraid to even imagine what will happen next! - surprise, fear \\
\hspace*{0.5cm} I can't believe someone would eat that! It's horrible!" - disgust, surprise \\
\\\\
\textbf{No emotions / other emotions}
\\\\
\textbf{No emotions}
\\
\hspace*{0.5cm} This morning it was raining. \\
\hspace*{0.5cm} He read the book in two days. \\
\hspace*{0.5cm} 
I saw her yesterday on the street.
\\\\
\textbf{Other emotions}
 \\
\hspace*{0.5cm} I am extremely annoyed with this mess! \\
\hspace*{0.5cm} My heart is breaking with pain :(  \\
\hspace*{0.5cm} We finally did it :):) I'm just over the moon! 

\end{tcolorbox}

\begin{tcolorbox}[breakable, colback=blue!10!white, colframe=blue!50!black, title=\texttt{Main Instructions for the Second Project: Joy, Sadness, Anger}]
\selectlanguage{english}

Select one or more emotions and their intensity in the text. If there are no emotions in the text or if there are emotions not represented in the list, select the No emotions / other emotions option.
\\\\
Example
\\\\
\textbf{Joy}
\\
Low \\
\hspace*{0.5cm} Your smile makes my day.
\\\\
Normal \\
\hspace*{0.5cm} This is one of the best gifts I have ever received. \\ 
\hspace*{0.5cm} It was very fun and wonderful, our vacation was a success!!! \\
\\\\
High \\
\hspace*{0.5cm} We finally did it!!!!! I'm just over the moon \\
\hspace*{0.5cm} We won!!! :):) I can't believe it happened! \\
\\\\
\textbf{Sadness}
\\
Low \\
\hspace*{0.5cm} It was a hard day for me. \\
\\\\
Normal\\
\hspace*{0.5cm} I can't believe this happened to us...
\\\\
High\\
\hspace*{0.5cm} My heart is breaking with pain :((
\\\\
\textbf{Anger}
\\
Low \\
\hspace*{0.5cm} It pisses me off
\\\\
Normal \\
\hspace*{0.5cm} I am extremely annoyed with this mess!
\\\\
High \\
\hspace*{0.5cm} This is absolutely unacceptable!!! 
\\\\
\textbf{Examples with multiple emotions }\\
\hspace*{0.5cm} We finally found the perfect place to stay, and it's even better than I could have imagined! - joy, surprise \\
\hspace*{0.5cm} Wow, how unexpected, this is the best gift I've ever received! - joy, surprise \\
\hspace*{0.5cm} I'm glad you came, but you're so damn late! - joy, anger \\
\hspace*{0.5cm} It's hard for me to accept that it ended this way, and I'm angry with you for it. - sadness, anger \\
\hspace*{0.5cm} It's so sad and depressing that our relationship ended because of your lies! - anger, sadness 
\\\\
\\\\
\textbf{No emotions / other emotions} \\
\hspace*{0.5cm} This morning it was raining. \\
\hspace*{0.5cm} He read the book in two days. \\
\hspace*{0.5cm} I saw her yesterday on the street.

\end{tcolorbox}

\section{Labels Co-occurrence Statistics}
\label{sec:app_labels_cooccurrence}
\begin{figure*}[h!]
    \centering
    \includegraphics[width=\linewidth]{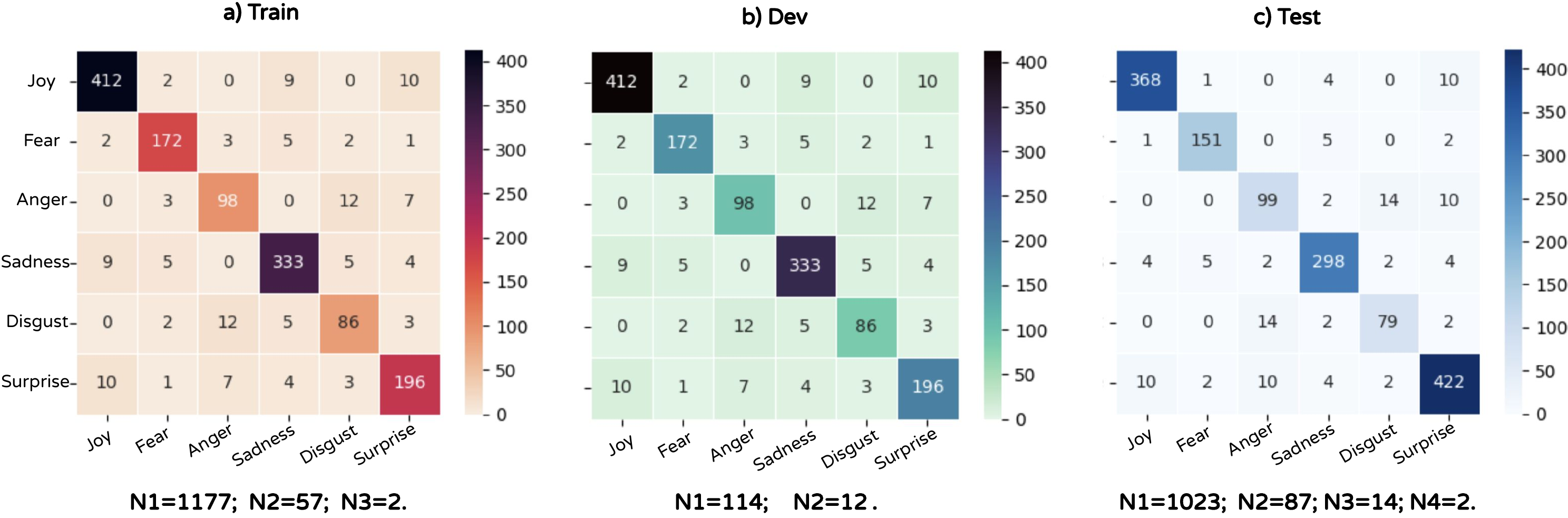}
    \caption{Labels co-occurrence statistics.}
    \label{fig:labels_cooccurrence}
\end{figure*}

Additionally to the overall train, development, and test splits, we also report emotion co-occurrence within these splits, as shown in Figure~\ref{fig:labels_cooccurrence}. The majority of texts are labeled with a single emotion. However, approximately 6\% of texts in the dataset are annotated with two or more emotions. Among the most frequent co-occurrences are \texttt{Joy} with \texttt{Surprise} and \texttt{Disgust} with \texttt{Anger}, reflecting natural patterns of emotional expression, though other combinations are also observed. A promising future direction for this work is the annotation of more fine-grained and diverse emotional texts, potentially supported by semi-automated methods using our released baseline model.

\newpage

\section{\emo Samples Examples}
\label{sec:app_samples}

\begin{table}[ht!]
    \centering
    \begin{tabular}{p{1.75cm}|p{11cm}|c}
    \toprule
        \multicolumn{1}{c|}{\textbf{Emotion}} & \multicolumn{1}{c|}{\textbf{Data Examples}} & \textbf{Intensity}\\
        \midrule
        \rowcolor{LightOrange} & \foreignlanguage{ukrainian}{То так мило і гарно.} \newline \textcolor{gray}{\scriptsize{\textit{It's so nice and beautiful.}}} & \textsc{Low} \\
        \rowcolor{LightOrange} \textsc{Joy} & \foreignlanguage{ukrainian}{вже майже час слухаю співи, це справді шикарно*-*} \newline \textcolor{gray}{\scriptsize{\textit{I've been listening to the singing for almost an hour now, it's really great*-*}}} & \textsc{Medium} \\
        \rowcolor{LightOrange} & \foreignlanguage{ukrainian}{І найголовніше, з Новим роком, пташки!!!} \newline \textcolor{gray}{\scriptsize{\textit{And most importantly, Happy New Year, birds!!!}}} & \textsc{High} \\
        
        \midrule
        \rowcolor{LightPurple} & \foreignlanguage{ukrainian}{Бо я прокинулась, глянула в дзеркало і злякалась.} \newline \textcolor{gray}{\scriptsize{\textit{Because I woke up, looked in the mirror, and got scared.}}} & \textsc{Low} \\
        \rowcolor{LightPurple} \textsc{Fear} & \foreignlanguage{ukrainian}{Поспала годинку і почали снитись жахіття :(} \newline \textcolor{gray}{\scriptsize{\textit{I slept for an hour and started having nightmares :(}}} & \textsc{Medium} \\
        \rowcolor{LightPurple} & \foreignlanguage{ukrainian}{А в мене руки трусяться) !!!} \newline \textcolor{gray}{\scriptsize{\textit{And my hands are shaking) !!!}}} & \textsc{High} \\
        
        \midrule
        \rowcolor{LightRed} & \foreignlanguage{ukrainian}{Спілкувалась я з деякими, і от бісить і всьо тут} \newline \textcolor{gray}{\scriptsize{\textit{I talked to some of them, and this is what makes me angry}}} & \textsc{Low} \\
        \rowcolor{LightRed} \textsc{Anger} & \foreignlanguage{ukrainian}{Ставте крапку, мати вашу я знав!} \newline \textcolor{gray}{\scriptsize{\textit{Put a stop to it, I knew your mother, damn it!}}} & \textsc{Medium} \\
        \rowcolor{LightRed} & \foreignlanguage{ukrainian}{Просто чоооорт, ну якого я такий ідіот?!?!} \newline \textcolor{gray}{\scriptsize{\textit{Why am I such an idiot?!?!}}} & \textsc{High} \\
        
        \midrule
        \rowcolor{LightBlue} & \foreignlanguage{ukrainian}{Але за дітками і їхніми обнімашки скучила.} \newline \textcolor{gray}{\scriptsize{\textit{But I missed my children and their hugs.}}} & \textsc{Low} \\
        \rowcolor{LightBlue} \textsc{Sadness} & \foreignlanguage{ukrainian}{Не виходить смачний чай:// вкотре} \newline \textcolor{gray}{\scriptsize{\textit{I can't make delicious tea:// once again}}} & \textsc{Medium} \\
        \rowcolor{LightBlue} & \foreignlanguage{ukrainian}{Але вона не живе зі мною (((( і я сумую.} \newline \textcolor{gray}{\scriptsize{\textit{But she doesn't live with me (((( and I miss her.}}} & \textsc{High} \\
        
        \midrule
        \rowcolor{LightGreen} & \foreignlanguage{ukrainian}{В Києві душно, брудно, нудно і нема чим дихати.} \newline \textcolor{gray}{\scriptsize{\textit{Kyiv is stuffy, dirty, boring, and there is no air to breathe.}}} & \textsc{Low} \\
        \rowcolor{LightGreen} \textsc{Disgust} & \foreignlanguage{ukrainian}{Гірлянди там галімі, а свічки смердючі.} \newline \textcolor{gray}{\scriptsize{\textit{The lights are crappy, and the candles are stinky.}}} & \textsc{Medium} \\
        \rowcolor{LightGreen} & \foreignlanguage{ukrainian}{відповідь очевидна – там лайно, фууу!!} \newline \textcolor{gray}{\scriptsize{\textit{the answer is obvious - it's shit, ewww!}}} & \textsc{High} \\
        
        \midrule
        \rowcolor{LightYellow} & \foreignlanguage{ukrainian}{не може бути, а чому?} \newline \textcolor{gray}{\scriptsize{\textit{it can't be, and why?}}} & \textsc{Low} \\
        \rowcolor{LightYellow} \textsc{Surprise} & \foreignlanguage{ukrainian}{а шо це, шоце? я шось не бачила такого?} \newline \textcolor{gray}{\scriptsize{\textit{what's this, what's this? I haven't seen anything like it?}}} & \textsc{Medium} \\
        \rowcolor{LightYellow} & \foreignlanguage{ukrainian}{а я то думала...он воно що!!} \newline \textcolor{gray}{\scriptsize{\textit{and here I was thinking... but that's it!!!}}} & \textsc{High} \\
        
        \midrule
         & \foreignlanguage{ukrainian}{Знову вертоліт над \#lviv} \newline \textcolor{gray}{\scriptsize{\textit{Helicopter over \#lviv again}}} & \\
         \textsc{None} & \foreignlanguage{ukrainian}{поки що не хочу дітей} \newline \textcolor{gray}{\scriptsize{\textit{i don't want children yet}}} & \\
         & \foreignlanguage{ukrainian}{Гуляю собі галицьким селом тихою дорогою.} \newline \textcolor{gray}{\scriptsize{\textit{I'm walking along a quiet road in one Halychyna village.}}} & \\
    \bottomrule
    \end{tabular}
    \caption{\emo dataset examples per each emotions.}
    \label{tab:app_examples}
\end{table}

\newpage

\section{LLMs Prompts for Emotions Classification}
\label{sec:app_prompts}

Here, we provide exact prompts used for LLMs prompting for emotion classification task in Ukrianian texts. We used two types of prompts: instructions in English and instructions in Ukrianian.

\begin{tcolorbox}[breakable, colback=black!5!white,        
  colframe=gray!10!black,      
  width=\linewidth,       
  boxrule=0.4mm,          
  arc=2mm,                
  outer arc=2mm,          
  boxsep=2.5mm,             
  title=\texttt{Prompt with Instructions in English}]
  
Evaluate whether the following text conveys any of the following emotions: joy, fear, anger, sadness, disgust, surprise.
\\
If the text does not have any emotion, answer neutral. \\
One text can have multiple emotions. \\
Think step by step before you answer. Answer only with the name of the emotions, separated by comma.
\\\\
Examples:\\\\
Text: \foreignlanguage{ukrainian}{Але, божечко, як добре вдома.} \\
Answer: joy
\\\\
Text: \foreignlanguage{ukrainian}{Я в п'ятницю признавалась в коханні  і мене відшили!}\\
Answer: sadness
\\\\
Text: \foreignlanguage{ukrainian}{Починаю серйозно хвилюватись за котика.} \\
Answer: fear
\\\\
Text: \foreignlanguage{ukrainian}{Я тебе ненавиджу, п'яна як може бути!} \\
Answer: anger
\\\\
Text: \foreignlanguage{ukrainian}{Тут смердить і мальчіки з синім волоссям п'ють.}\\
Answer: disgust
\\\\
Text: \foreignlanguage{ukrainian}{А що, цей канал досі існує?} \\
Answer: surprise
\\\\
Text: \foreignlanguage{ukrainian}{Хочу вже наводити порядок в новому домі.} \\
Answer: neutral
\\\\
Text: {input} \\
Answer:
\end{tcolorbox}

\begin{tcolorbox}[breakable, colback=black!5!white,        
  colframe=gray!10!black,      
  width=\linewidth,       
  boxrule=0.4mm,          
  arc=2mm,                
  outer arc=2mm,          
  boxsep=2.5mm,             
  title=\texttt{Prompt with Instructions in Ukrainian}]
\selectlanguage{ukrainian}

Оціни, чи передає текст будь-які з цих емоцій: радість, злість, страх, сум, здивування, огида. \\
Якщо в тексті немає емоцій, відповідай нейтральна. \\
Один текст може викликати багато емоцій. \\
Думай крок за кроком, перш ніж відповідати. Відповідай тільки назвами емоцій розділених комою.
\\\\
Приклади:
\\\\
Тект: Але, божечко, як добре вдома. \\
Відповідь: радість
\\\\
Тект: Я в п'ятницю признавалась в коханні  і мене відшили!\\
Відповідь: сум
\\\\
Тект: Починаю серйозно хвилюватись за котика.\\
Відповідь: страх
\\\\
Тект: Я тебе ненавиджу, п'яна як може бути!\\
Відповідь: злість
\\\\
Тект: Тут смердить і мальчіки з синім волоссям п'ють.\\
Відповідь: огида
\\\\
Тект: А що, цей канал досі існує?\\
Відповідь: здивування
\\\\
Тект: Хочу вже наводити порядок в новому домі.\\
Відповідь: нейтральна
\\\\
Текст: {input} \\
Відповідь:
\end{tcolorbox}  

\section{Model hyperparameters}
\label{sec:app_hyperparameters}

Here, we report the hyperparameters details for the utilized models. 

Table 5
reports the tuned \texttt{learning rates} per each Transformer-encoder based models. Within all models, we used \texttt{batch size} 64, 50 \texttt{epochs} with \texttt{early stopping callback} 3 according to the accuracy of the evaluation. Many models stopped their training steps at $10^{\text{th}}$-$15^{\text{th}}$ epoch.

For LLMs, for generation, we used default hyperparameters per model with no additional changes.

\begingroup
\renewcommand{\arraystretch}{1.15}
\begin{table}[h!]
\centering
\footnotesize
\begin{tabular}{c|c}
\toprule
\textbf{Model} & \textbf{Learn. rate} \\
\toprule
LaBSE & 1E-04 \\ \hline
Geotrend-BERT & 1E-04 \\ \hline
mBERT & 1E-05 \\ \hline
UKR-RoBERTa Base & 1E-05 \\ \hline
XLM-RoBERTa Base & 1E-05 \\ \hline
XLM-RoBERTa Large & 1E-05 \\ \hline
Twitter-XLM-RoBERTa & 1E-04 \\ \hline
Glot500 Base & 1E-06 \\ \hline
Multilingual-E5 Large & 1E-05 \\ \hline
Multilingual-E5 Base & 1E-05 \\
\bottomrule
\end{tabular}
\label{tab:hyperparameters}
\caption{The best learning rate for the Transformer-encoder based models fine-tuned on original Ukrainian data.}
\end{table}
\endgroup

\newpage

\section{Confusion Matrices}
\label{sec:app_conf_matrices}

Here, in addition to the main results, we also report the confusion matrices for the top performing models.

\begin{figure}[ht!]
    \centering
     \includegraphics[scale=0.5]{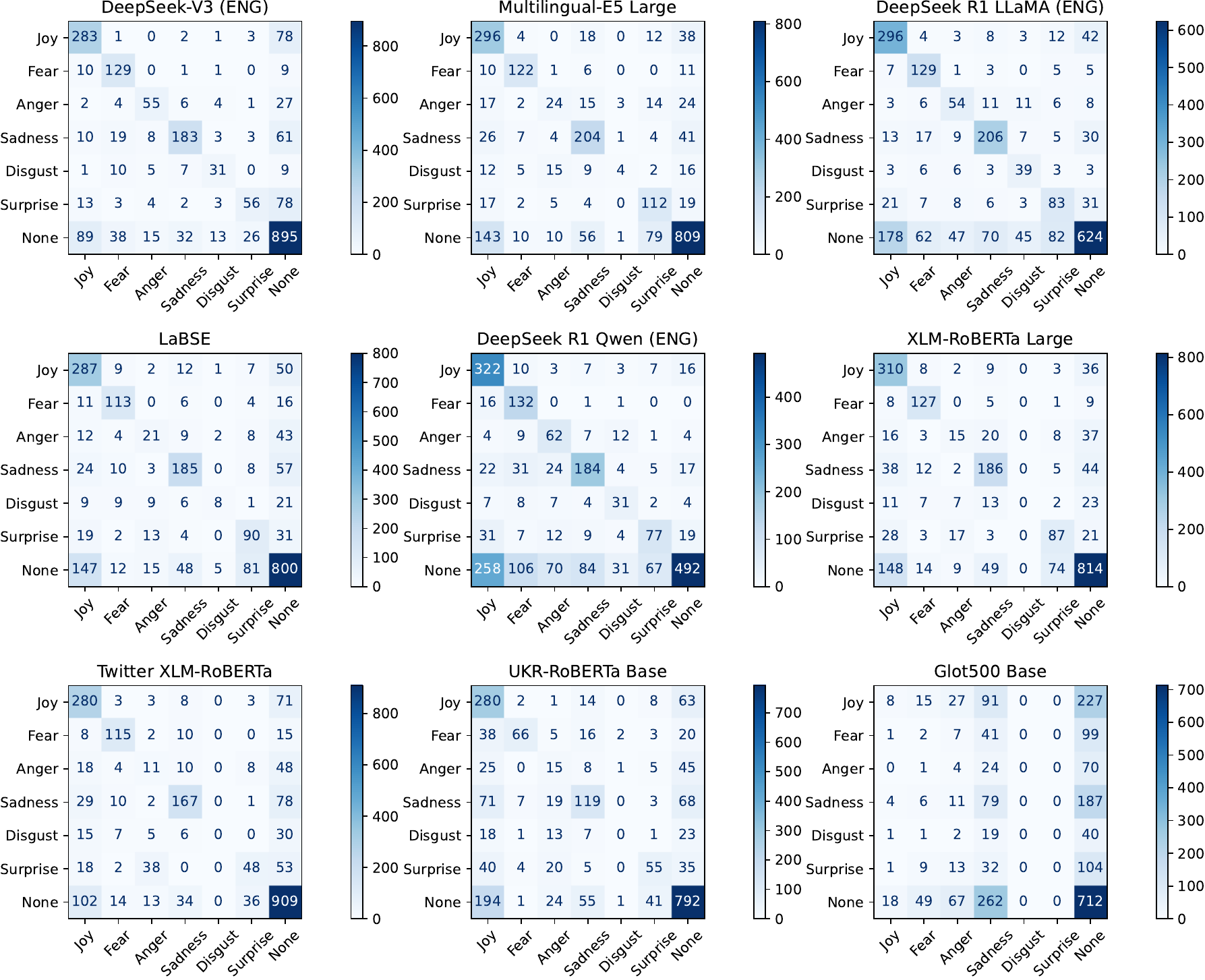}
    \caption{Confusion matrices of the top performing models fine-tuned on the \emo training data.}
    \label{fig:informed_consent}
    \end{figure}

\end{document}